\documentclass[onecolumn, conference]{IEEEtran}

\title{AttEntropy: On the Generalization Ability of Supervised Semantic Segmentation Transformers to New Objects in New Domains}

\makeatletter % changes the catcode of @ to 11
\newcommand{\linebreakand}{%
  \end{@IEEEauthorhalign}
  \hfill\mbox{}\par
  \mbox{}\hfill\begin{@IEEEauthorhalign}
}
\makeatother     

\author{
\IEEEauthorblockN{Krzysztof Lis${}^\dagger$}
\IEEEauthorblockA{
 lis.krzysztof@protonmail.com}
 \and
 \IEEEauthorblockN{Matthias Rottmann${}^{\star\dagger}$}
\IEEEauthorblockA{rottmann@uni-wuppertal.de}
\and
\IEEEauthorblockN{Annika Mütze${}^\star$}
\IEEEauthorblockA{
muetze@uni-wuppertal.de}
\linebreakand
\IEEEauthorblockN{Sina Honari${}^\ddagger$}
\IEEEauthorblockA{ sina.honari@gmail.com}
\and
 \IEEEauthorblockN{Pascal Fua${}^\dagger$}
\IEEEauthorblockA{
 pascal.fua@epfl.ch}
 \and
\IEEEauthorblockN{Mathieu Salzmann${}^\dagger$}
\IEEEauthorblockA{ mathieu.salzmann@epfl.ch}
\linebreakand
\IEEEauthorblockN{${}^\dagger$
\textit{Computer Vision Laboratory, EPFL, Lausanne, Switzerland}
}
\linebreakand
\IEEEauthorblockA{${}^\star$ \textit{IZMD \& School of Mathematics and Natural Sciences}, \textit{University of Wuppertal}, \textit{Wuppertal, Germany}
}
\linebreakand
\IEEEauthorblockA{${}^\ddagger$ \textit{ Samsung AI Center Toronto,}  
\textit{ Toronto, Canada}
}
}

\usepackage[backend=bibtex, style=numeric]{biblatex}
\usepackage{graphicx}
\usepackage{multirow}
\usepackage{multicol}
\usepackage{booktabs} %
\usepackage{hyperref}

\usepackage{amsmath}
\usepackage{amssymb}

\usepackage[capitalize]{cleveref}

\usepackage{pifont} %

\usepackage{colortbl}
\definecolor{Gray}{gray}{0.88}
\definecolor{DarkGray}{gray}{0.78}
\newcolumntype{h}{>{\columncolor{DarkGray}}c}
\newcolumntype{g}{>{\columncolor{Gray}}c}

\usepackage{import}
\usepackage{wrapfig}
\usepackage[normalem]{ulem}

\newcommand{\citeViT}{Dosovitskiy20}
\newcommand{\citeSETR}{Zheng21}
\newcommand{\citeSegformer}{Xie21b}

\newif\ifdraft

\newcommand{\segmetricsA}[1]{
	\multicolumn{2}{c}{Pixel-level} 
	& \multicolumn{3}{#1}{Segment-level}
}

\newcommand{\segmetricsB}{
	AP $\uparrow$ 
	& FPR$_{95}$ $\downarrow$ 
	& $\overline{\mbox{sIoU}}\uparrow$ 
	& $\overline{\mbox{PPV}}\uparrow$ 
	& $\overline{F_1}\uparrow$
}

\newcommand{\segmetricsShort}{
	AP $\uparrow$ 
	& FPR$_{95}$ $\downarrow$ 
}

\newcommand{\svgwidth}{\textwidth}

\begin{document}

\maketitle

\begin{abstract}
In addition to impressive performance, vision transformers have demonstrated remarkable abilities to encode information they were not trained to extract. For example, this information can be used to perform segmentation or single-view depth estimation even though the networks were only trained for image recognition. 
We show that a similar phenomenon occurs when \emph{explicitly} training transformers for semantic segmentation in a \emph{supervised} manner for a set of categories: Once trained, they provide valuable information even about categories {\it absent} from the training set. This information can be used to segment objects from these never-seen-before classes in domains as varied as road obstacles, aircraft parked at a terminal, lunar rocks, and maritime hazards.
\end{abstract}

%

%
%

% !TEX root = ../top.tex
% !TEX spellcheck = en-US
    
\definecolor{FigOrange}{RGB}{200, 50, 10}
\definecolor{FigBlue}{RGB}{50, 50, 100}

\begin{figure} [ht]
{\setlength{\tabcolsep}{2pt}
\renewcommand{\arraystretch}{0.5}
    \centering
    \begin{tabular}{cc}
    \includegraphics[height=0.180\linewidth]{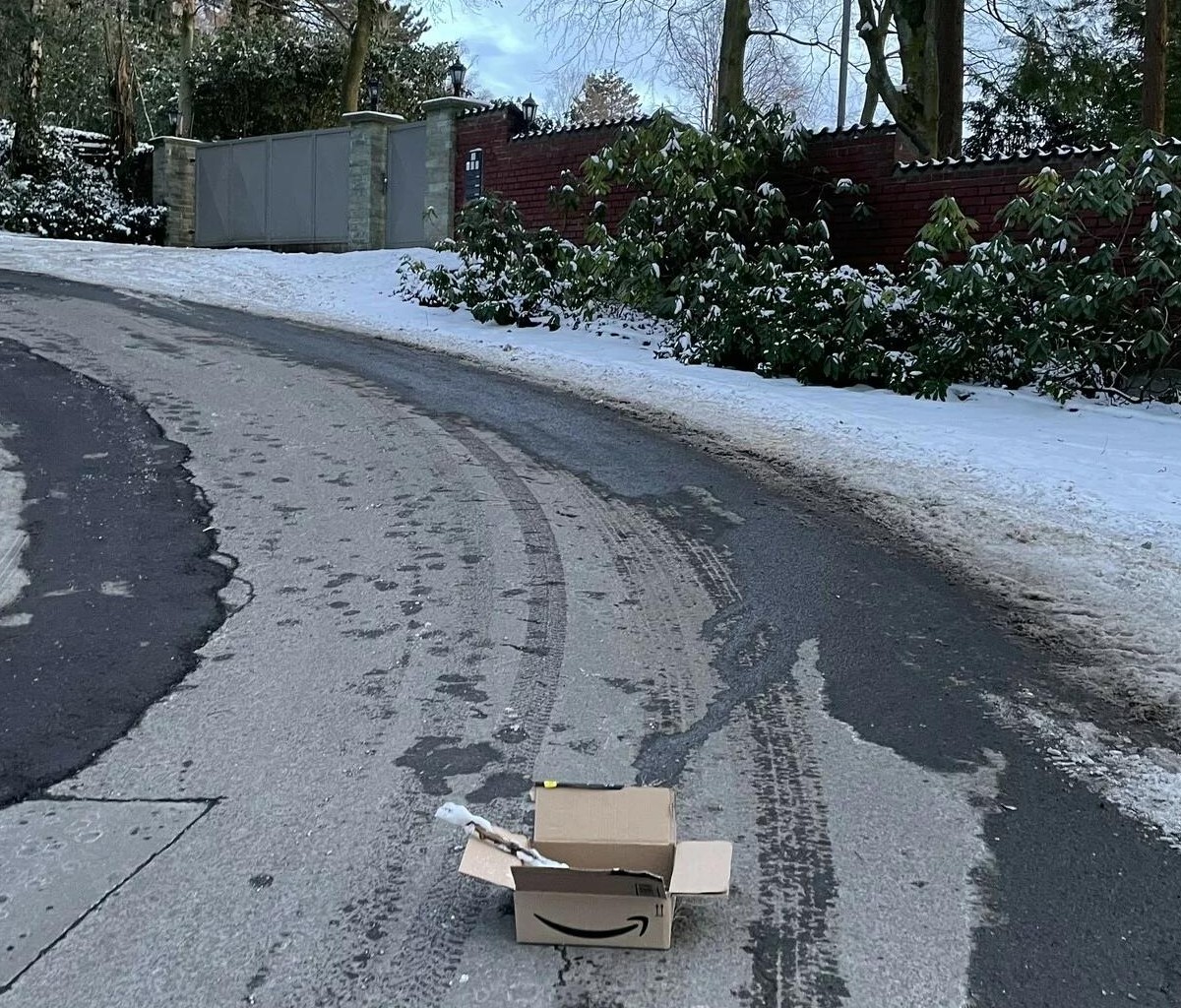}
    \includegraphics[height=0.180\linewidth]{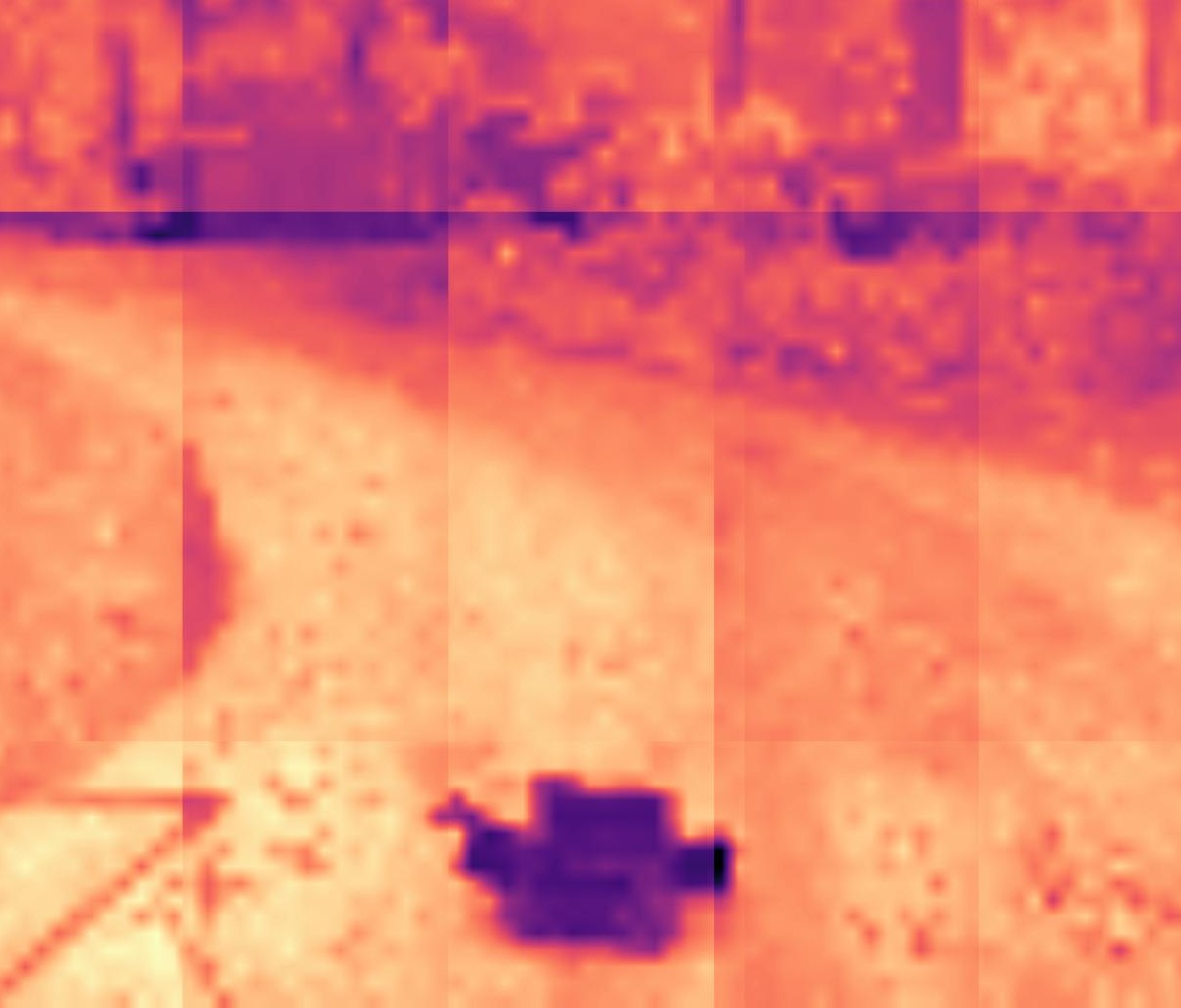}%
    &
    \includegraphics[height=0.185\linewidth]{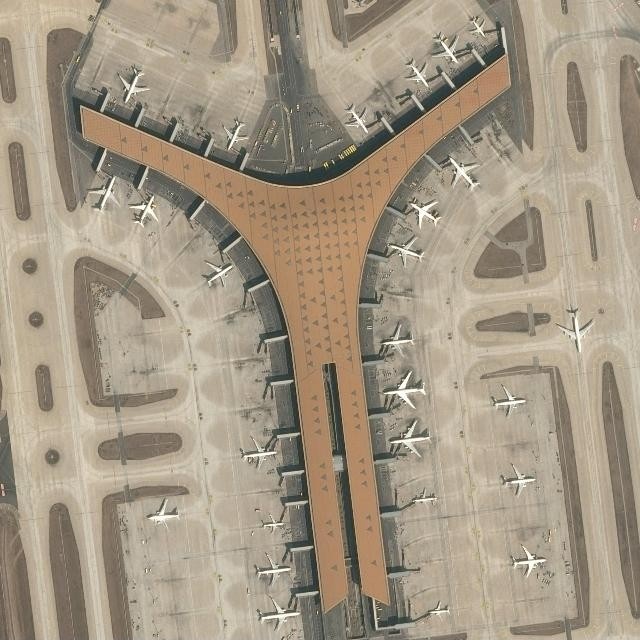}
    \includegraphics[height=0.185\linewidth]{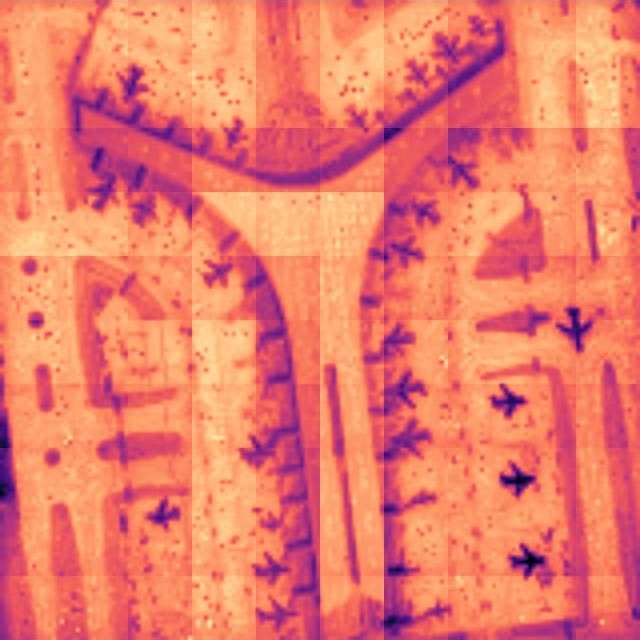}%
    \\%
    % {\small Unusual road obstacles\\\cite{Chan21b}} & {\small Aircrafts\\ \cite{AirbusAircraftSegmentation}} \\
    {\footnotesize Unusual road obstacles } & {\footnotesize Aircraft }%
    \\ %
    \includegraphics[height=0.185\linewidth]{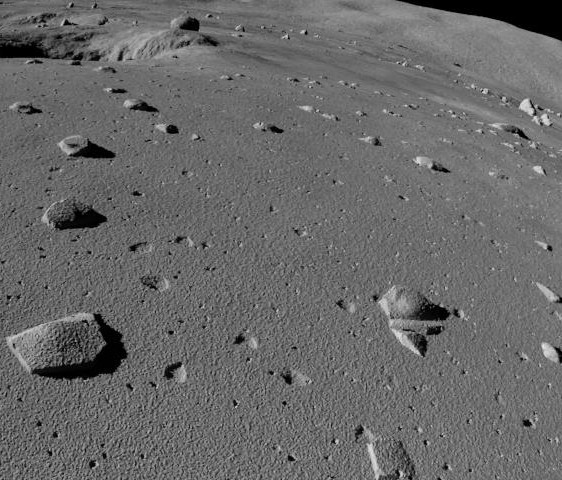}
    \includegraphics[height=0.185\linewidth]{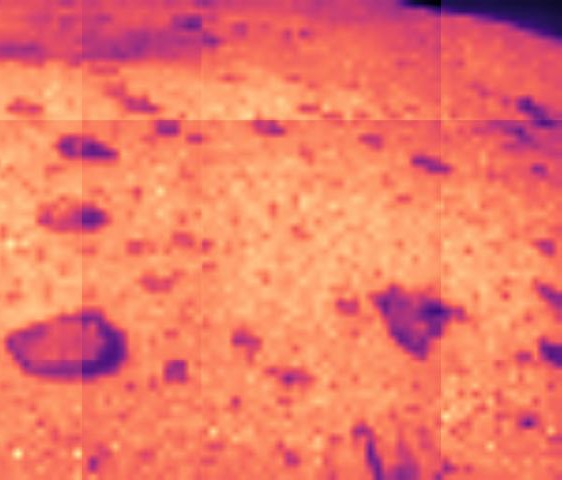}
    &
    \includegraphics[height=0.185\linewidth]{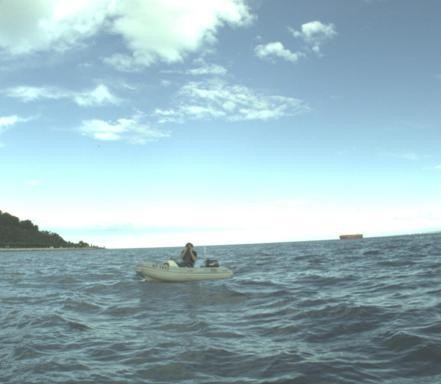}
    \includegraphics[height=0.185\linewidth]{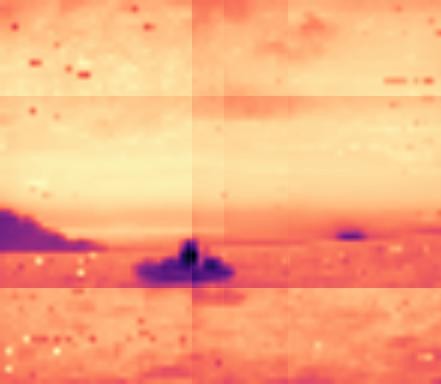}% 
    \\%
    {\footnotesize Lunar rocks} & {\footnotesize Maritime hazards}%
    % Lunar rocks\\\cite{ArtificialLunarLandscape} & Maritime hazards\\\cite{Bovcon19}
    \end{tabular}
    \vspace{1pt}
    \caption{{\bf Segmenting objects from categories the network was {\it not} trained for.}
    The attention-entropy extracted from a SETR~\cite{\citeSETR} transformer trained on the urban driving Cityscapes~\cite{Cordts16} dataset lets us segment objects from four previously unseen categories. In all four cases, we show an original image (left) and the corresponding Attention Entropy (right).}
    \label{fig:qualiDomainTour}
}
\end{figure}

\section{Introduction}
\label{sec:intro}
Vision transformers have become increasingly prominent in the computer vision community. They can handle many tasks from image recognition~\cite{\citeViT} to semantic segmentation~\cite{\citeSETR,Ranftl21dpt,\citeSegformer,Strudel2021_Segmenter,Cheng2021_MaskFormer} and single-view depth estimation~\cite{chang2021transformer,monovit}. Furthermore, they exhibit many remarkable properties. For example when trained for image recognition, they can be used to segment the main object in the input image~\cite{strudel2021segmenter}. Similarly self-supervised vision transformers can be used to segment the foreground objects using the attention of the class-token~\cite{caron2021emerging} or by partitioning a graph of inter-patch affinity~\cite{LOST,wang2022tokencut}. This is evidence that even transformers \emph{not} explicitly trained for segmentation encode information relevant to it.

In this paper, we show that a similar phenomenon occurs when \emph{explicitly} training transformers for semantic segmentation in a \emph{supervised} manner for a given set of categories: Once trained, they provide valuable information about object categories {\it absent}  the training set and this information can be used to segment objects from these never-seen-before classes. \Cref{fig:qualiDomainTour} illustrates this behavior for objects from very different categories. To this end, we consider the intermediate spatial attention maps of segmentation transformers and extract the Shannon entropy of spatial attentions. We refer to it as {\it AttEntropy} and show that it provides the information we require. We show experimentally that when a transformer is trained using {\it any} sufficiently large training set -- Cityscapes~\cite{Cordts16}, ImageNet~\cite{Deng09}, or ADE20k~\cite{ade20k} -- AttEntropy is sufficiently informative to allow segmentations for objects as different as road obstacles, aircraft parked at a terminal~\cite{AirbusAircraftSegmentation}, lunar rocks~\cite{ArtificialLunarLandscape}, and maritime hazards~\cite{Bovcon19}.  To further demonstrate the practicality of our approach, we evaluate it quantitatively on road obstacle segmentation, a key task for self-driving for which many suitable benchmarks are available~\cite{Chan21b,pinggera2016lost}. We show that it does particularly well on  small to moderate-size objects, which are not addressed by approaches mostly designed to find large foreground items~\cite{caron2021emerging,LOST,wang2022tokencut}. 
Our contributions can be summarized as follows:
\begin{itemize}
    \item We study the behavior of vision transformers trained in a supervised fashion for semantic segmentation when faced with new objects and new domains.
	\item We extract the entropy of spatial attentions and show that it can be used to segment objects from previously unseen categories.
	\item We demonstrate the generality of our analysis using three different transformer architectures~\cite{\citeSETR,\citeSegformer,Ranftl21dpt} trained on three different datasets.
\end{itemize}
Our code is publicly available at \url{https://github.com/adynathos/AttEntropy}.

\section{Related Work}\label{sec:related}

\textbf{Segmentation / Localization and Attention.} 
A few works use features or attention from transformer backbones for object detection~\cite{gupta2022ow,LOST} or semantic segmentation~\cite{athanasiadis2022weaklysupervised,hamilton2022unsupervised,caron2021emerging,wang2022tokencut}.
In particular, the methods of~\cite{athanasiadis2022weaklysupervised,hamilton2022unsupervised} rely on transformer features for object segmentation.
{However, the focus of these methods is not the segmentation of}
unseen objects, but rather on {the segmentation of objects}
present in the database under weak supervision~\cite{athanasiadis2022weaklysupervised} or even without supervision~\cite{hamilton2022unsupervised}.

Closest to our work, are those of~\cite{caron2021emerging,wang2022tokencut,LOST,melas2022deep} that rely on transformers trained in a self-supervised manner. They can be considered to some extent as operating in an open world setting as the transformers have been trained on ImageNet~\cite{Deng09} and then applied to other datasets.
The work of~\cite{caron2021emerging} performs segmentation based on the final layer class tokens from the transformer attention.
The works of~\cite{wang2022tokencut,melas2022deep} use the transformer of~\cite{caron2021emerging}. They then constructs a graph based on the last layer features, and performs segmentation via a graph cut algorithm. Unlike in our approach, the attention itself is not utilized in~\cite{wang2022tokencut}. 
Likewise, the method of~\cite{LOST} computes the similarity between patches using key-features of the last attention layer, selects a seed patch and segments a group of patches similar to it, which yields a single foreground object.
{In addition, a similarity between retrieval-based obstacle segmentation and self-attention modules of transformers was shown in the training-free algorithm ROAS~\cite{fu2023evolving}. However, they exploit priors specific for driving scenarios and use features of convolutional neural networks whereas we study the information encoded in the spatial attention maps of transformers.}
In contrast to the works~\cite{wang2022tokencut,caron2021emerging,LOST,fu2023evolving}, we consider the attention between image patches across different layers, study their properties and use an information theoretic approach to segment new objects in new domains. More importantly,
our work constitutes the first study showing that a semantic segmentation transformer trained in a supervised fashion on known classes inherently learns to segment unknown objects, irrespective of the given context.

\textbf{Obstacle Detection.} 
To show the practicability of the spatial attention property, we benchmark on obstacle detection in road and maritime scenes. To this end, we briefly review existing work in that field. Most of the road obstacle segmentation methods and the closely related road anomaly segmentation ones can be grouped into two main classes: 1) Pasting synthetically generated obstacles or obstacles from other datasets into an image and then learning to detect them~\cite{bevandic2022dense,grcic2021dense,grcic2022densehybrid,Lis22, grcic2023advantages,grcic2024dense}, and 2) performing in-painting, sometimes conditioned on predicted segmentation masks, and then assessing the deviation from the input~\cite{DiBlase21,Lis19,Xia20,Creusot15,Lis20}.
Besides these approaches, proxy datasets for anomalies or collections of prediction errors have also been used to learn uncertainties on anomalies and obstacles~\cite{chan2020entropy,Oberdiek_2020_CVPR_Workshops,Bruegge20}. {In addition, decoupling classification and mask learning allows to learn outlier-aware segmentation by exploiting the structural correlation of nearby pixels~\cite{zhang2024csl,rai2023unmasking} and specific outlier scoring functions~\cite{grcic2023advantages,nayal2023rba}.}

Methods designed to learn the abstract notion of road obstacle or of road anomaly are rare. 
Most of them rely on uncertainty estimation, having been designed for image classification and being modified only in minor ways~\cite{gal2016dropout,hendrycks2019scaling}. 
They usually serve as baselines.

\section{From Attention to Segmentation}\label{sec:method}

We first remind the reader of the attention mechanism that is at the heart of all transformer architectures. We then show how to use the attention matrices to create entropy heatmaps that highlight small to moderate-size objects. Finally, we discuss our implementation. 

\subsection{Attention Mechanism in Vision Transformers} 

\definecolor{FigOrange}{RGB}{200, 50, 10}
\definecolor{FigBlue}{RGB}{50, 50, 100}

\begin{figure} 
\centering
{\scriptsize
\renewcommand{\svgwidth}{\textwidth}
%% Creator: Inkscape inkscape 0.92.5, www.inkscape.org
%% Accompanies image file 'attention_maps.pdf' (pdf, eps, ps)
\begingroup%
  \makeatletter%
  \providecommand\color[2][]{%
    \errmessage{(Inkscape) Color is used for the text in Inkscape, but the package 'color.sty' is not loaded}%
    \renewcommand\color[2][]{}%
  }%
  \providecommand\transparent[1]{%
    \errmessage{(Inkscape) Transparency is used (non-zero) for the text in Inkscape, but the package 'transparent.sty' is not loaded}%
    \renewcommand\transparent[1]{}%
  }%
  \providecommand\rotatebox[2]{#2}%
  \newcommand*\fsize{\dimexpr\f@size pt\relax}%
  \newcommand*\lineheight[1]{\fontsize{\fsize}{#1\fsize}\selectfont}%
  \ifx\svgwidth\undefined%
    \setlength{\unitlength}{783.75bp}%
    \ifx\svgscale\undefined%
      \relax%
    \else%
      \setlength{\unitlength}{\unitlength * \real{\svgscale}}%
    \fi%
  \else%
    \setlength{\unitlength}{\svgwidth}%
  \fi%
  \global\let\svgwidth\undefined%
  \global\let\svgscale\undefined%
  \makeatother%
  \begin{picture}(1,0.28948408)%
    \lineheight{1}%
    \setlength\tabcolsep{0pt}%
    \put(0,0){\includegraphics[width=\unitlength,page=1]{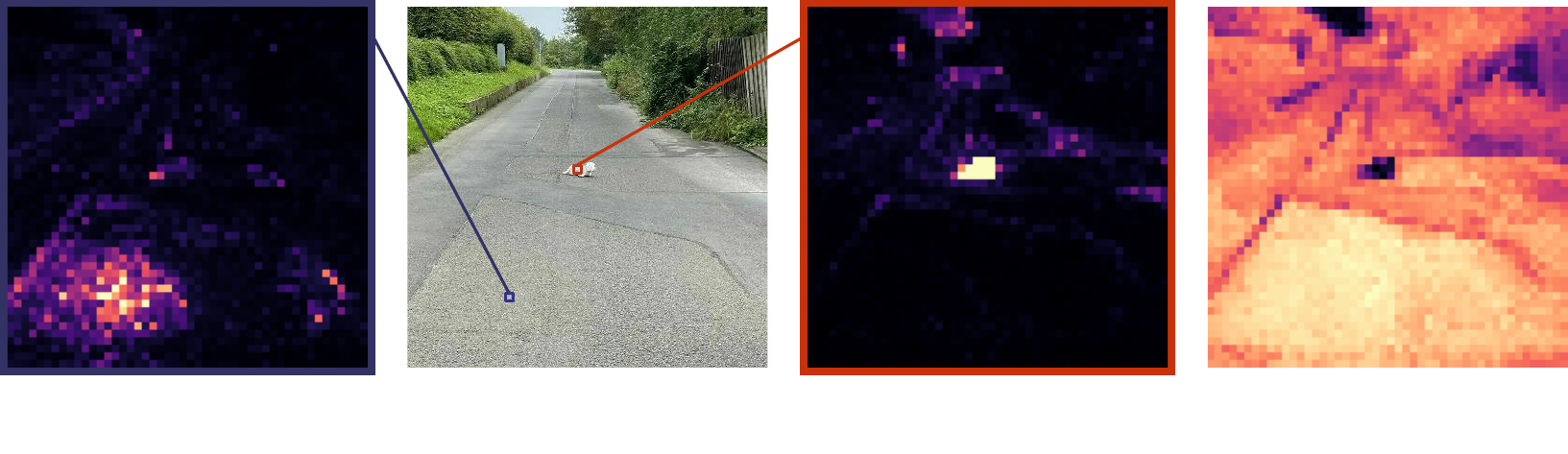}}%
    \put(0.11996936,0.03486825){\color[rgb]{0,0,0}\makebox(0,0)[t]{\lineheight{1.25}\smash{\begin{tabular}[t]{c}Attention map\\ of a road patch\end{tabular}}}}%
    \put(0.63033614,0.03486825){\color[rgb]{0,0,0}\makebox(0,0)[t]{\lineheight{1.25}\smash{\begin{tabular}[t]{c}Attention map \\of an obstacle patch\end{tabular}}}}%
    \put(0.88468521,0.03349027){\color[rgb]{0,0,0}\makebox(0,0)[t]{\lineheight{1.25}\smash{\begin{tabular}[t]{c}Entropy \\of attention distributions\end{tabular}}}}%
    \put(-0.20574164,0.54785722){\color[rgb]{0,0,0}\makebox(0,0)[lt]{\begin{minipage}{1.41626794\unitlength}\centering \end{minipage}}}%
  \end{picture}%
\endgroup%

}%
\vspace{-25pt}
\caption{{\bf Entropy heatmaps.} For each image patch, we compute the Shannon entropy of outgoing attentions. We show the spatial attention at two different image locations. The right image shows the Shannon entropy. Small objects receive concentrated attention and thus low corresponding entropy. An interactive tool for attention visualization is included in the supplementary material (and will be publicly available).}
\label{fig:attention_maps}
\end{figure}
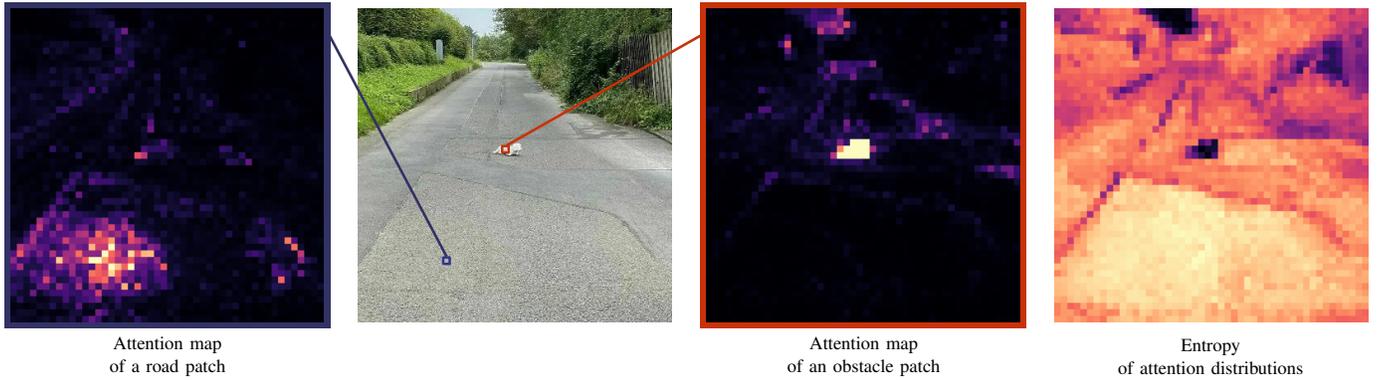

The Vision Transformer (ViT)~\cite{\citeViT} is one of the first successful applications of the transformer self-attention mechanism to image inputs. With enough training data, it can be more powerful than a traditional convolutional network for image classification.
We base the following description on the popular ViT architecture, but will show in the supplement how it can be implemented for the differently structured Segformer~\cite{\citeSegformer}.
Furthermore, it serves as a backbone feature extractor for the Segmentation Transformer (SETR)~\cite{\citeSETR}, providing improvement in semantic segmentation of road scenes over comparable convolutional backbones.

ViT first decomposes the image in a checkerboard fashion into $N \times N$ image patches, each of size $16 \times 16$ pixels, calculates their initial encodings and adds a learned spatial embedding, which will allow nearby patches to attend strongly to each other.
The patches serve as tokens in the transformer. The attention mechanism connects all the patches to each other.

The backbone consists of $L$ multi-head self-attention ($\mathit{MSA}$) blocks. 
Each attention block, or layer, $l$ receives as input a triplet 
$(\mbox{query}\;Q,\; \mbox{key}\;K,\; \mbox{value}\;V)$
computed from an input feature map $Z^{l-1}\in \mathbb{R}^{N^2 \times C}$, where $C$ denotes the number of 
feature
channels. The triplet is computed from $Z^{l-1}$ as
\begin{equation}
	\mathit{Q} = Z^{l-1} W_Q, ~~
	\mathit{K} = Z^{l-1} W_K, ~~ 
	\mathit{V} = Z^{l-1} W_V,
\end{equation}
where $W_Q, W_K, W_V \in \mathbb{R}^{C \times d}$ are learnable weight matrices and $d \in \mathbb{N}$ a parameter which we specify below. The attention $A^l$, which is the matrix that concerns us most in this paper,  is taken to be
\begin{equation}
	A^l(Z^{l-1}) = \mathrm{softmax}( Q \cdot K^T / \sqrt{d} ) \, .
\end{equation}
The softmax acts row-wise, so that the outgoing attention of each token sums up to 1.
In the multi-head scenario, the attention is in fact computed multiple times. This yields $m$ independent attention matrices $\mathit{A}^l_i$, $i=1,\ldots,m$.
{As in \cite{\citeSETR}, we choose $d = C/m$. While we will later focus on the attention matrices, for the sake of completeness the recursive calculation of $Z^{l}$ is described in the supplementary material.}

\textbf{Properties of the Attention Tensors.} 
As described above, ViT~\cite{\citeViT} computes attention matrices $A^{l}_{i}(Z^{l-1})$ for layers $l=1,\ldots,L$ and for multiple heads $i=1,\ldots,m$. Due to the softmax activation and the decomposition of the image into patches, these attentions can be viewed as 
probability distributions over the $N \times N$ image patches plus the extra class-token, which we discard due to its lack of a geometric interpretation. For all $j , j' \in \{1,\ldots,N^2\}$ 
in layer $l$, the element $A^{l}_{i}(Z^{l-1})_{j,j'}$ of the attention matrix reflects how much attention patch $j$ pays to patch $j'$. 
These elements can therefore be visualized as heatmaps over the $N\times N$ image patches, as shown in Fig.~\ref{fig:attention_maps},
where we visualize the attention averaged over the multiple heads
\begin{equation}
\bar{A}^{l}(Z^{l-1})_{j,j'} = \frac{1}{m} \sum_{i=1}^m A^{l}_{i}(Z^{l-1})_{j,j'} \, .
\end{equation}
Attention is driven by visual similarity and, thanks to the spatial embedding, proximity.
We observe that the attention originating from a visually distinct, moderately sized object
remains concentrated sharply within the patches $j$ overlapping to a greater extent with that object.
By contrast, for larger areas of visual coherence, such as the road in a traffic scene, the attention is more dispersed over the entire region.

\subsection{From Attention to Entropy Heatmaps}

We quantify this behavior by estimating the \emph{Spatial Shannon Entropy}
\begin{equation}
	E^l(Z^{l-1})_j =  - \sum_{j'=1}^{N^2} \bar{A}^{l}(Z^{l-1})_{j,j'} \cdot \log( \bar{A^{l}}(Z^{l-1})_{j,j'} )
\end{equation}
for each image patch $j=1,\ldots,N^2$. The overall spatial entropy $\mathbf{E}^l = [ E^l(Z^{l-1})_j ]_{j \in \{1,\ldots,N^2\}}$ for layer $l$ can be viewed as an aggregated heatmap for layer $l$ that can be used to segment objects of moderate size, as discussed below.

\textbf{Layer Selection and Averaging.}
Each self-attention layer yields its own entropy heatmap $\mathbf{E}^l$.
Inspecting the entropy heatmaps reveals that objects, such as the obstacles in Fig.~\ref{fig:attention_maps},
usually have low entropy, due to their concentrated attention.
Some isolated background patches, such as parts of the road in Fig.~\ref{fig:attention_maps}, nonetheless exhibit lower or higher entropy than their neighbors.
Averaging entropy across layers tends to suppress these artifacts.

This behaviour is exhibited to varying degrees by different layers.
The object segmentation capability can be further improved by selecting a subset $\mathcal{L} \subseteq \{1,\ldots,l\}$ of layer entropy heatmaps and averaging, computing $
    \mathbf{\bar{E}} = \frac{1}{L} \sum_{l \in \mathcal{L}} \mathbf{E}^l \, .$
We devise a simple strategy to select layers $\mathcal{L}$ which are best suited for object segmentation.
We compute the entropy values for a synthetic input image containing a circle, a stand in for a generic object, on a textured background.
We then select the layers which exhibit meaningfully lower entropy in the circle than in the background.
We study the impact of this strategy and provide a detailed visualization of per-layer attentions and entropies in the supplementary material.

\textbf{Segmentation of Entropy Heatmaps.}
The averaged entropy is used as an object detection heatmap with minimal postprocessing.
First, we negate the entropy since objects exhibit {\it lower} entropy than their surroundings.
Then we linearly interpolate the entropy signal from the original resolution of $N \times N$ to a per pixel heatmap in the image resolution and apply a thresholding to it. This corresponds to
\begin{equation}
    s(u,v) = \delta (\text{Lin}_{k \in \text{Neighbour(u,v)}}(-\mathbf{\bar{E}}_k)) \,,
\end{equation}
where $s(u,v)$ is the binary segmentation value for pixel location $(u,v)$, $\delta$ is a threshold function for binarization, $\text{Lin}$ is the linear interpolation applied to the neighbouring elements $k$ that contribute to pixel $(u, v).$
The chosen threshold for $\delta$, which is applied to the segmentation heatmap, makes a trade-off between precision and recall. In our quantitative evaluations, we report the AP over all possible thresholds
following the evaluation protocol of~\cite{Chan21b}.

\iffalse

The given transformer backbone comprises $L=24$ layers. We observed that intermediate backbone layers, i.e., $l=7,\ldots,13$, exhibit this behavior the most. Towards later layers of the backbone, the attention seems to focus more on semantics while the early layers close to the image input focus on texture and the intermediate layers on visual features. We present supporting image material in the appendix. 

\textbf{Spatial Shannon Entropy.}
%
The concentration of the attention $\bar{A}^{l}(Z^{l-1})$ whenever there is an objects with a small to intermediate size located in patch $i$, motivates the usage of the Shannon entropy to quantify the peakedness of the spatial distribution $\bar{A}^{l}(Z^{l-1})$. The more peaked $\bar{A}^{l}(Z^{l-1})$ is on an object of limited spatial extent, the smaller the entropy. The more $\bar{A}^{l}(Z^{l-1})$ disperses, e.g.\over the whole street, the higher the entropy. Hence we compute the spatial Shannon entropy for each patch$j \in \{1,\ldots,N^2\}$ via 
\begin{equation}
	E^l(Z^{l-1})_j =  - \sum_{j'}^{N^2} \bar{A}^{l}(Z^{l-1})_{j,j'}
	\cdot \log( \bar{A^{l}}(Z^{l-1})_{j,j'} ) \, .
\end{equation}

\fi

%
%

\section{Experiments}\label{sec:experiments}

In our experiments, we extract attention entropy from Visual Transformer (ViT)~\cite{\citeViT} and Segformer~\cite{\citeSegformer} backbones
and observe the entropy to be a useful cue for object segmentation in both cases.
We use the backbones of SETR~\cite{\citeSETR} trained with Cityscapes, DPT~\cite{Ranftl21dpt} trained with ADE20k, and Segformer~\cite{\citeSegformer} trained with Cityscapes.
We demonstrate that the attention layers trained in a supervised fashion to segment specific categories carry the required information to segment new objects of small to moderate size under varying degrees of domain shift and finally when changing the domain completely. 
We do so by evaluating our method and several baselines on a variety of datasets. 
We show that spatial attention information is much more powerful than a number of widely used training-free strategies, and that 
semantic segmentation training is particularly beneficial for this usage of attention.

\subsection{Datasets}\label{sec:exp_datasets}

In our study, we use segmentation transformers trained on Cityscapes~\cite{Cordts16} due to their availability and because there exist multiple datasets containing obstacles that are semantically different from the objects present in Cityscapes. 
We then take these pre-trained transformers and evaluate them on the following datasets to detect new objects.

\textbf{Segment Me If You Can~\cite{Chan21b}}
is a benchmarking framework for the segmentation of traffic anomalies and obstacles previously unseen during training.
We exploit its {\it Road Obstacles 21} track, a dataset of diverse and small obstacles located within a variety of road scenes, 
including difficult weather and low light.
The ground-truth labels are kept private so no training can be performed on the test set.
We follow the benchmark's evaluation protocol and metrics, which measure the accuracy of classifying pixels as belonging to obstacles or the road surface, as well as object-level detection measures.

\textbf{Lost and Found~\cite{pinggera2016lost}}
also features previously unseen obstacles on road surfaces, with a region of interest constrained to the drivable space.
The background streets and parkings are relatively similar to that of the Cityscapes training data.
The {\it test no known} is the testing subset used in the {\it Segment Me} benchmark by excluding Cityscapes classes.

\textbf{MaSTr1325~\cite{Bovcon19} (Maritime)} 
contains images of maritime scenes viewed from a small unmanned surface vessel
with semantic segmentation labels for {\it water, sky} and {\it obstacles}.

\textbf{Artificial Lunar Landscape~\cite{ArtificialLunarLandscape} (Lunar)}
comprises synthetic lunar scenes where rocks act as obstacles.

\textbf{Airbus Aircraft Segmentation~\cite{AirbusAircraftSegmentation} (Aircraft)} features satellite images of airports.

\noindent 
Our networks are not trained on these datasets. Instead we use publicly available Cityscapes and ADE20k checkpoints. 
On the latter 3 datasets, we show the models can detect unseen objects in completely new domains.

\subsection{Gradually Changing the Domain}

\begin{table*}[t]
\centering
\setlength{\heavyrulewidth}{1pt}
\setlength{\tabcolsep}{4pt}
{

\resizebox{\textwidth}{!}{
    \begin{tabular}{ll|rrrrr|rrrrrlll}
    \toprule
    {} & {} & \multicolumn{5}{c|}{Lost and Found - test no known} & \multicolumn{5}{c}{Road Obstacles 21 - test} \\
    {} & {} & \segmetricsA{c|} & \segmetricsA{c} \\
    {} & {} & \segmetricsB & \segmetricsB \\
    \midrule
    
    \multirow{14}{*}{\shortstack{No \\ training}}
    {} & Ensemble \cite{Lakshminarayanan17} & 2.9 & 82.0 & 6.7 & 7.6 & 2.7 & 1.1 & 77.2 & 8.6 & 4.7 & 1.3 \\
    {} & Embedding Density \cite{Blum19} & 61.7 & 10.4 & 37.8 & 35.2 & 27.5 & 0.8 & 46.4 & 35.6 & 2.9 & 2.3  \\
    {} & LOST \cite{LOST} &1.1 & 94.7 & 8.6 & 6.0 & 6.0 & 4.7 & 93.8 &17.0 & 8.3 &11.0 \\
    {} & MC Dropout \cite{Mukhoti18} & 36.8 & 35.5 & 17.4 & 34.7 & 13.0 & 4.9 & 50.3 & 5.5 & 5.8 & 1.0 \\
    {} & Max Softmax \cite{Hendrycks17b} & 30.1 & 33.2 & 14.2 & 62.2 & 10.3 & 15.7 & 16.6 & 19.7 & 15.9 & 6.3 \\
    {} & Mahalanobis\cite{Lee18a} & 55.0 & 12.9 & 33.8 & 31.7 & 22.1 & 20.9 & 13.1 & 13.5 & 21.8 & 4.7 \\
    {} & ODIN \cite{Liang18b} & 52.9 & 30.0 & 39.8 & 49.3 & 34.5 & 22.1 & 15.3 & 21.6 & 18.5 & 9.4 \\
    {} & DINO \cite{caron2021emerging} & 26.4 & 38.9 &11.7 &13.6 & 5.7 & 39.9 & 14.1 &26.8 &19.1 &12.4 \\
    {} & M2F-EAM \cite{grcic2023advantages}& - & - & - & - & - & {66.9} & 17.9 & - & - & -\\
    {} & RbA \cite{nayal2023rba}& - & - & - & - & - & {\bf 87.8} & 3.3 & 47.4 & 56.2 & {\bf 50.4}\\
    {} & RAOS (trainig free) \cite{fu2023evolving} & {\bf 81.1} & 3.2 & 53.5 & 46.2 & {\bf 50.6} & \underline{79.7} & 0.8 & 42.4 & 32.6 & 32.7\\
    % {} & Ours Segformer$_{all}$ & 34.7 & 11.9 & 22.2 & 35.7 & 19.9 & 47.4 & 15.7 & 29.7 & 36.4 & 25.2 \\
    % {} & Ours DPT$_{all}$ & 42.6 & 44.5 & 16.0 & 48.6 & 20.1 & 32.7 & 28.0 & 11.3 & 25.6 & 12.0 \\
    
    % {} & Ours Segformer$_{manual}$ & 56.4 & 6.7 & 34.7 & 35.0 & 28.4 & 45.5 & 8.1 & 25.4 & 36.3 & 22.7 \\
    {} & Ours Segformer & 51.2 & 10.8 & 37.8 & 35.6 & 28.9 & 32.1 & 11.5 & 24.3 & 30.1 & 18.6 \\
    {} & Ours DPT & 57.0 & 5.7 & 17.2 & 31.8 & 18.3 & 55.1 & 3.8 & 20.8 & 42.8 & 24.1 \\
    % {} & Ours DPT$_{manual}$ & 64.6 & 2.5 & 25.0 & 48.2 & 28.4 & 61.4 & 5.5 & 20.0 & 32.6 & 21.1 \\
    % {} & Ours SETR$_{all}$ & 67.0 & 5.8 & 30.6 & 22.1 & 22.2 & 62.2 & 11.2 & 27.4 & 28.2 & 24.9 \
    % {} & Ours SETR$_{manual}$ & {\bf 73.0} & 2.9 & 37.1 & 42.8 & {\bf 38.0} & {\bf 72.9} & 2.5 & 36.4 & 47.8 & {\bf 41.6} \\
    {} & Ours SETR & \underline{73.0} & 3.9 & 35.5 & 43.9 & \underline{37.5} & 71.3 & 2.5 & 35.7 & 46.3 & \underline{39.7} \\
    \bottomrule
    \end{tabular}
}	
}
% \vspace{2pt}
\caption{\textbf{Obstacle detection scores}. 
The primary metrics are Average Precision (AP$\uparrow$) for pixel classification 
and Average $F_1$ ($\overline{F_1}\uparrow$) for instance level detection. The evaluation protocol and metrics follow the {\it Segment Me If You Can} benchmark~\cite{Chan21b}. 
}
\label{table:benchmarks_baselines}
\end{table*}

\begin{figure*} 
\centering
\includegraphics[width=0.21\linewidth]{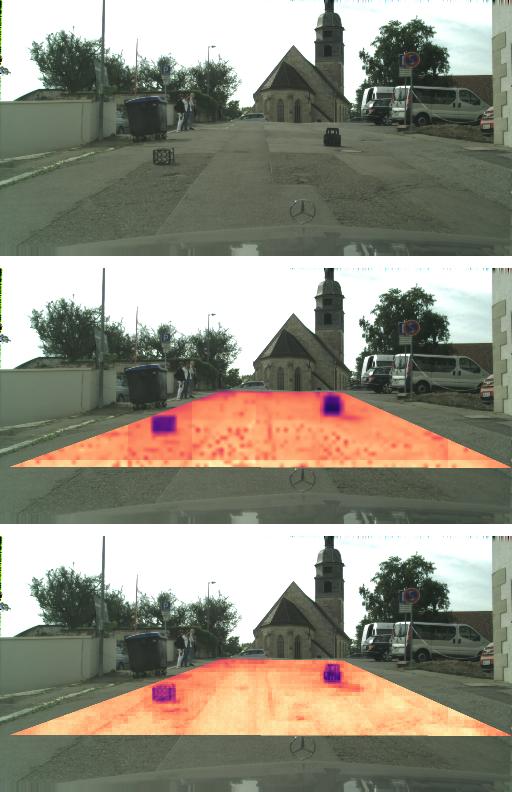}
\includegraphics[width=0.21\linewidth]{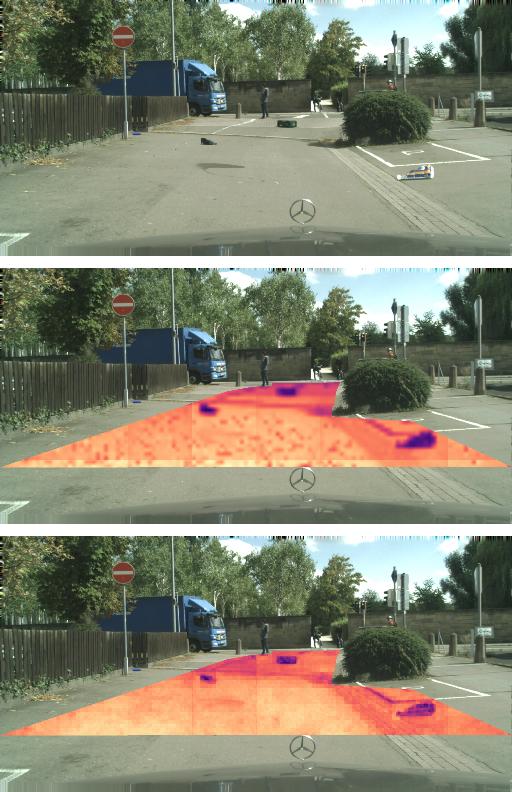}
\includegraphics[width=0.185\linewidth]{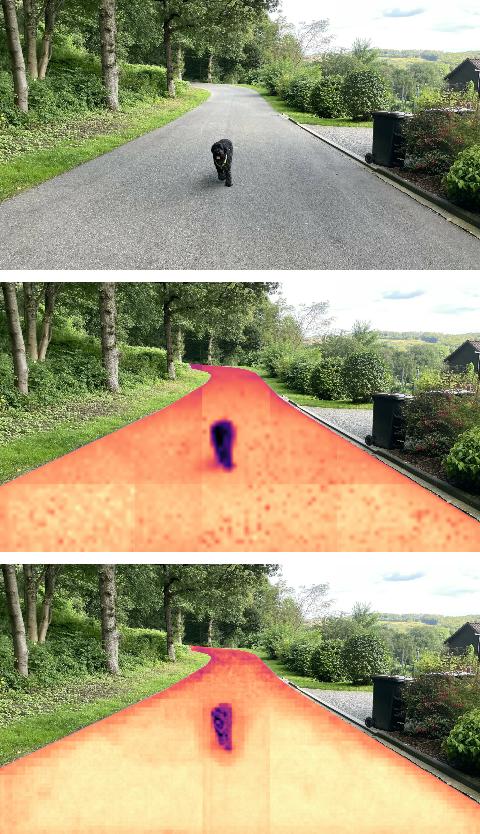}
\includegraphics[width=0.185\linewidth]{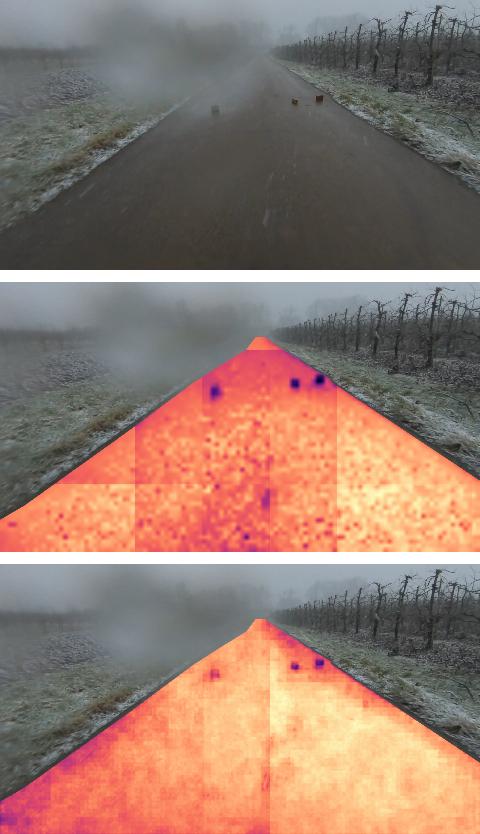}
\includegraphics[width=0.185\linewidth]{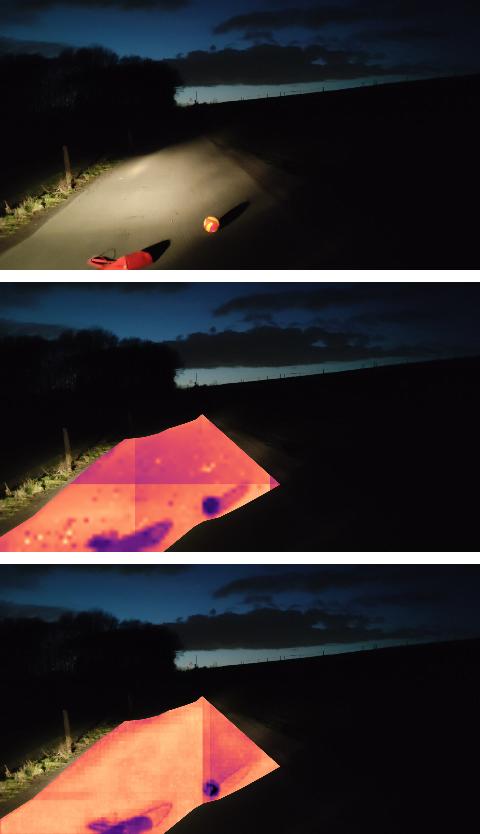}
% \includegraphics[width=0.24\linewidth]{images/quali_segme/04_Maurener_Weg_8_000008_000140.jpg}
% \includegraphics[width=0.24\linewidth]{images/quali_segme/07_Festplatz_Flugfeld_000001_000220.jpg}
%jpg
\caption{
Qualitative results on obstacle detection in traffic scenes. The left two images are from LostAndFound~\cite{Pinggera16} and the right three images are from {RoadObstacle21}~\cite{Chan21b}, including ones with difficult weather and limited light.
The middle and bottom rows show the averaged entropy of SETR and Segformer respectively, both using manual layer averaging.
The heatmap is overlaid in the evaluation ROI.
%Qualitative results on obstacle detection in traffic scenes with difficult weather and limited light.
%The middle and right columns show the averaged entropy of SETR and Segformer respectively, both using manual layer averaging.
Slight rectangular artifacts arise from MMSegmentation's sliding window inference.
}
\label{fig:quali_segme}
\end{figure*}

% \begin{figure*} 
% \centering
% %
% \includegraphics[width=0.24\linewidth]{images/quali_segme/validation_40.jpg}
% \includegraphics[width=0.24\linewidth]{images/quali_segme/curvy-street_carton_6.jpg}
% \includegraphics[width=0.24\linewidth]{images/quali_segme/04_Maurener_Weg_8_000008_000140.jpg}
% \includegraphics[width=0.24\linewidth]{images/quali_segme/07_Festplatz_Flugfeld_000001_000220.jpg}

% \includegraphics[width=0.24\linewidth]{images/quali_segme/validation_40_setr.jpg}
% \includegraphics[width=0.24\linewidth]{images/quali_segme/curvy-street_carton_6_setr.jpg}
% \includegraphics[width=0.24\linewidth]{images/quali_segme/04_Maurener_Weg_8_000008_000140_setr.jpg}
% \includegraphics[width=0.24\linewidth]{images/quali_segme/07_Festplatz_Flugfeld_000001_000220_setr.jpg}

% \includegraphics[width=0.24\linewidth]{images/quali_segme/validation_40_segf.jpg}
% \includegraphics[width=0.24\linewidth]{images/quali_segme/curvy-street_carton_6_segf.jpg}
% \includegraphics[width=0.24\linewidth]{images/quali_segme/04_Maurener_Weg_8_000008_000140_segf.jpg}
% \includegraphics[width=0.24\linewidth]{images/quali_segme/07_Festplatz_Flugfeld_000001_000220_segf.jpg}

% %jpg
% \caption{Qualitative results on obstacle detection in traffic scenes.
% The middle and bottom rows show the averaged entropy of SETR and Segformer respectively.
% The heatmap is overlaid in the evaluation ROI.
% }
% \label{fig:quali_segme}
% \end{figure*}

We start by evaluating road-segmentation models on unknown obstacles in similar scenes and then gradually make a domain shift by first evaluating on challenging and un-observed road environments and later on completely different domains such as maritime, lunar, and aerial scenes.

\textbf{No Domain Shift.} We start by examining the obstacle segmentation performance and report results on LostAndFound~\cite{Pinggera16}. While the objects are unseen by models the background is similar to Cityscapes road scenes. 
We report results in the left part of \cref{table:benchmarks_baselines}.
The table contains benchmark results of methods that do not train specifically for the detection of road obstacles. 
Monte Carlo (MC) dropout and ensembles are approximations to Bayesian inference, relying on multiple inferences. All the other methods in the table, as well as ours, are purely based on post processing of either the network's output or embedded features. In particular our method applied to the SETR model ranges as second best compared to all other training-free baselines with respect to both pixel-level and segment-level metrics.
In \cref{table:benchmarks_baselines} it can be observed that the Segformer~\cite{\citeSegformer} variant performs slightly worse. 
Possible causes are the coarse resolution of its attention maps, along with the fact that Segformer does not use the standard positional encoding found in most visual transformers.
With the positional codes influencing the query and key vectors, the tokens can easily learn to focus their attention on neighbors.
In lieu of such encoding, Segformer relies on $3 \times 3$ convolutions, interspersed with the attention layers, to leak positional information from zero-padding on the image edges.
Therefore, attention concentrations on very small objects may be less likely to emerge. 
Illustrative examples of the segmentation performance of our method are provided in \cref{fig:quali_segme}-left.
We have also observed that small obstacles close to the horizon can pose a problem to our method. Typically, the attention tends to concentrate in that region of the image as the street narrows in. We expect that digging deeper into the attention structure, as well as utilizing an auxiliary model supplied with our attention masks could yield a strongly performing overall system.

\textbf{Partial Domain Shift.} Next we evaluate models on RoadObstacle21~\cite{Chan21b}. While still being a road dataset, it provides unseen objects in new weather, lighting and background environments compared to Cityscapes. The observations from LostAndFound generalize to this dataset, see right-hand section of \cref{table:benchmarks_baselines}.
In the presence of slight to moderate domain shift, the performance of our method is preserved {and thereby ranges under the top three. In the supplementary material we provide an enlarged table with methods with obstacle or anomaly training. Our training-free AttEntropy reaches the same level as some training-based methods like SynBoost~\cite{DiBlase21}.} See visualization results in \cref{fig:quali_segme}-right, which further stress the capabilities of our method when weather and lighting conditions change drastically. 
In these visually difficult conditions, most of the obstacles present are still segmented well from the background.

\begingroup
\begin{figure} 
    \renewcommand{\arraystretch}{0.5}
    \begin{tabular}{p{0.505\textwidth}p{0.445\textwidth}}
        {\scriptsize
        \centering
        \newcommand{\svgwidth}{\linewidth}
        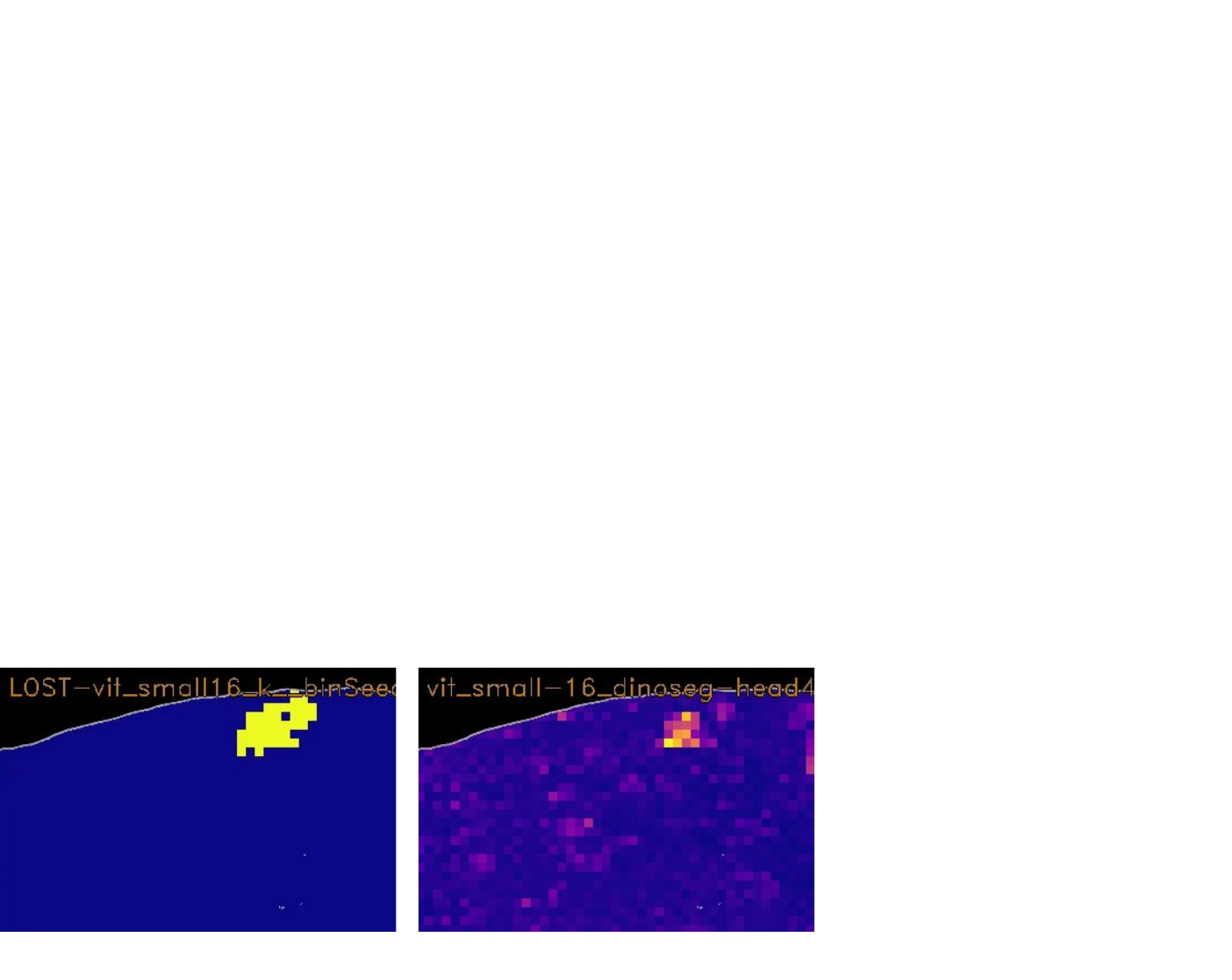
        }%
        % \includegraphics[width=0.30\linewidth]{images/sup_quali_water/0361_ours.jpg}
        % \includegraphics[width=0.30\linewidth]{images/sup_quali_water/0361_segf.jpg}
        % \makebox[0.3\linewidth]{\small Input} \makebox[0.3\linewidth]{\small Ours SETR}\makebox[0.3\linewidth]{\small Ours Segformer}
        
        % \includegraphics[width=0.30\linewidth]{images/sup_quali_water/0361_mahalanobis.jpg}%
        % \includegraphics[width=0.30\linewidth]{images/sup_quali_water/0361_odin.jpg}%
        % \includegraphics[width=0.30\linewidth]{images/sup_quali_water/0361_maxsoft.jpg}%
        % \makebox[0.3\linewidth]{\small Mahalanobis} \makebox[0.3\linewidth]{\small ODIN}\makebox[0.3\linewidth]{\small Max Softmax}
        
        % \includegraphics[width=0.30\linewidth]{images/sup_quali_water/0361_lost.jpg}%
        % \includegraphics[width=0.30\linewidth]{images/sup_quali_water/0361_dino.jpg}%
        % \includegraphics[width=0.30\linewidth]{images/sup_quali_water/0361_synboost.jpg}%
        
        %     \makebox[0.3\linewidth]{\small LOST} \makebox[0.3\linewidth]{\small DINO}\makebox[0.3\linewidth]{\small SynBoost}
        &
        {\scriptsize
        \newcommand{\svgwidth}{\linewidth}
        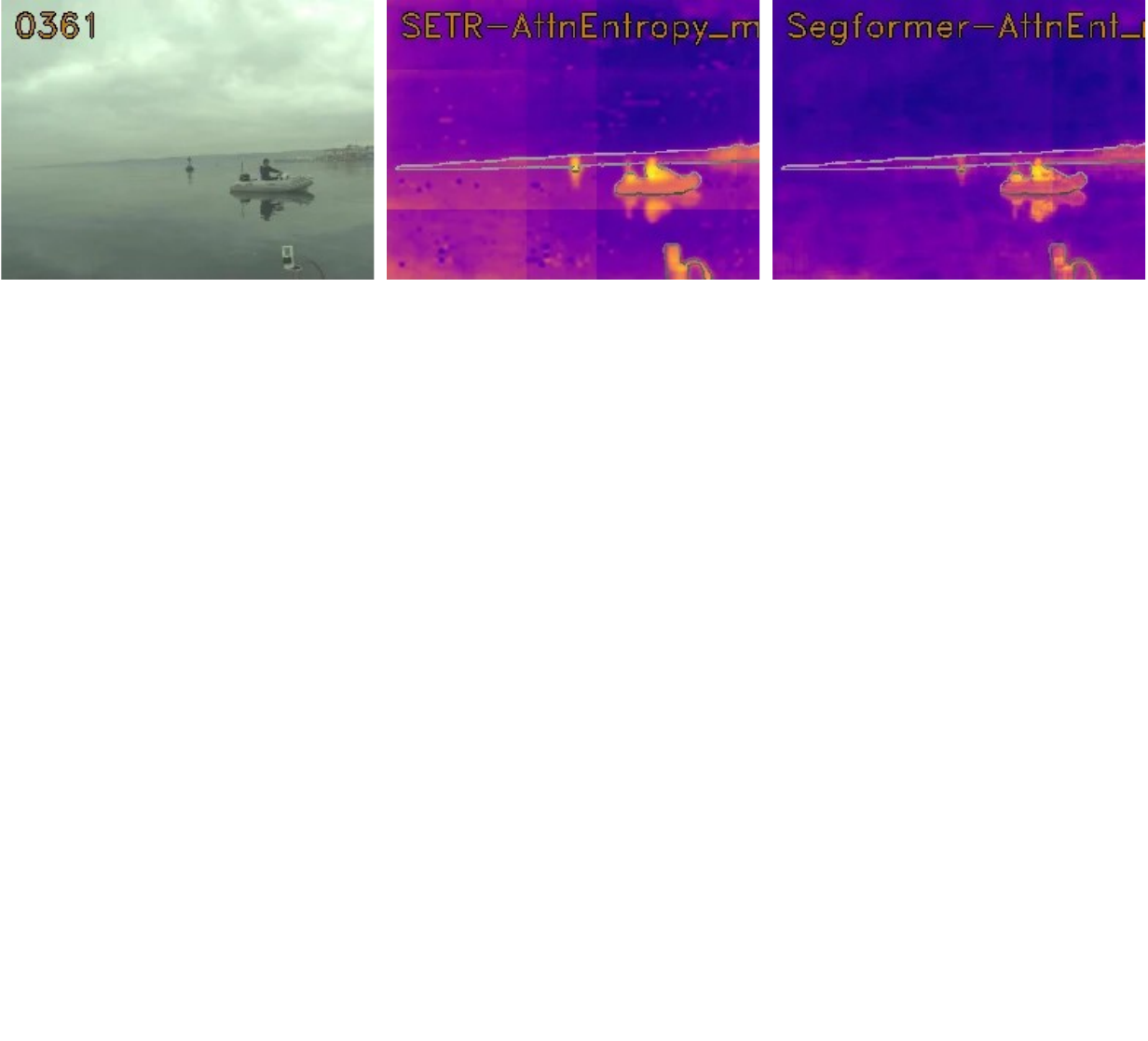
        }%
        \\
        \centering
        \vspace{-15pt}
         Lunar dataset~\cite{ArtificialLunarLandscape} &  \centering \vspace{-15pt} Maritime Dataset~\cite{Bovcon19} 
    \end{tabular}

        \caption{Qualitative results on obstacle detection on different datasets.
        The contour of the ground truth obstacle areas is highlighted.
        Compared to the other training-free obstacle detection methods, our attention entropy generalizes better to the distant domains.
        }
        \label{fig:sup_quali_lunar}
\end{figure}
\endgroup
\begin{table}[tb]
{
    \centering
    \setlength{\heavyrulewidth}{1pt}
    \resizebox{0.78\linewidth}{!}{
    {
    \begin{tabular}{l|rrrrr||rr}
    \toprule
    % {} & AP $\uparrow$ & FPR$_{95}$ $\downarrow$ \\
    {} & \multicolumn{5}{c||}{Artificial Lunar Landscape~\cite{ArtificialLunarLandscape}} & \multicolumn{2}{c}{MaSTr1325~\cite{Bovcon19}}\\
    {} & AP $\uparrow$ & FPR$_{95}$ $\downarrow$ & $\overline{\mbox{sIoU}}\uparrow$ & $\overline{\mbox{PPV}}\uparrow$ & $\overline{F_1}\uparrow$ &  AP $\uparrow$ & FPR$_{95}$ $\downarrow$ \\
    \midrule
    Ours SETR & 34.7 & 87.6 & {\bf 7.0} & 26.6 & {\bf 6.9}      & 59.6 & 39.6 \\
    Ours DPT & 35.1 & 89.9 & 4.0 & 37.7 & 5.1                &  \textbf{71.9} &       \textbf{14.2} \\
    Ours Segformer & 34.0 & 87.5 & 5.0 & 34.1 & 4.3           & 55.8 &      51.0 \\
    % Maximized Entropy \cite{Chan21a} & 41.5 & 81.0 & 5.0 & 43.5 & 6.0 \\
    % Ours Segformer$_{manual}$ & {\bf 37.8} & 76.1 & 5.6 & 38.2 & 5.1 \\
    % LOST-vit_small16_k__invDeg & 37.0 & 92.7 & 5.6 & 31.9 & 5.1 \\
    % Ours SETR & 35.9 & 85.0 & 7.9 & 29.8 & {\bf 8.3} \\
    SynBoost \cite{DiBlase21} & {\bf 35.6} & 81.9 & 4.0 & 28.8 & 3.6          & 14.9 & 80.0  \\
    DINO \cite{caron2021emerging} & 32.5 & 93.0 & 6.1 & 22.4 & 4.3            & 41.4 & 39.5 \\
    ODIN \cite{Liang18b} & 29.7 & {\bf 75.1} & 2.6 & 33.6 & 2.3                & 16.6 & 70.8 \\
    % Max Softmax \cite{Hendrycks17b} & 28.5 & 75.1 & 2.2 & 33.2 & 1.9 \\
    Max Softmax & 28.5 & {\bf 75.1} & 2.2 & 33.2 & 1.9                          & 18.6 & 64.9  \\
    LOST \cite{LOST} & 18.4 & 94.9 & 0.6 & {\bf 39.4} & 0.5                     & 10.1 & 94.4\\
    Mahalanobis \cite{Lee18a}& - & - & - & - &  -                                     & 5.9 & 97.3 \\
    \bottomrule
    \end{tabular}
    }} \vspace{10pt}
    \caption{\textbf{Obstacle pixel segmentation performance in the Lunar and Maritime dataset}.
    For the Lunar dataset the evaluation was performed on every 10-th image due to the big dataset size.
    The primary metrics are Average Precision (AP$\uparrow$) for pixel classification and Average $F_1$ ($\overline{F_1}\uparrow$) for instance level detection. For the Maritime dataset we do not measure object-level metrics since the maritime scenes contain non-instance obstacles, such as land.
    }%
    \label{table:lunar}%
    \label{table:maritime}%
}
% \parbox[b]{.43\linewidth}{
%     \centering
%     \setlength{\heavyrulewidth}{1pt}
%     % \setlength{\arrayrulewidth}{1pt}
%     \resizebox{0.75\linewidth}{!}{
%     \begin{tabular}{l|rr}
%     \toprule
%     {} & AP $\uparrow$ & FPR$_{95}$ $\downarrow$ \\
%     \midrule
%     Ours SETR  & 59.6 & 39.6 \\
%     Ours DPT &  \textbf{71.9} &       \textbf{14.2} \\
%     % Ours SETR$_{manual}$ & \textbf{63.9} & \textbf{35.0} \\
%     % LOST-vit\_small16\_k\_\_invDeg & 61.7 & 56.4 \\
%     Ours Segformer & 55.8 &      51.0 \\
%     % Maximized Entropy \cite{Chan21a} & 42.2 & 64.7 \\
%     DINO \cite{caron2021emerging} & 41.4 & 39.5 \\
%     Max Softmax  & 18.6 & 64.9 \\
%     ODIN \cite{Liang18b} & 16.6 & 70.8 \\
%     SynBoost \cite{DiBlase21} & 14.9 & 80.0 \\
%     LOST \cite{LOST} & 10.1 & 94.4 \\
%     Mahalanobis \cite{Lee18a} & 5.9 & 97.3 \\
%     \bottomrule
%     \end{tabular}
%     }	
%     \vspace{8pt}
%     \caption{\textbf{Obstacle pixel segmentation performance in the Maritime dataset MaSTr1325~\cite{Bovcon19}}.
%     We do not measure object-level metrics since the maritime scenes contain non-instance obstacles, such as land.
%     }
%     \label{table:maritime}
%     }
\end{table}

\textbf{Complete Domain Shift.}
We evaluate here the segmentation capability of models on Maritime, Lunar, and Aircraft datasets, which contain unseen objects in completely new domains compared to the road scenes.
The results for the Maritime dataset are presented in \cref{table:maritime}-right and the ones obtained from Lunar dataset are shown in \cref{table:lunar}-left. Among the methods we evaluated, our method outperforms the training-free baselines. The results show that not only can the attention entropy segment previously unseen road obstacles, but its small object detection property holds in completely different domains. 
On Maritime dataset, the objects on the water visually stand out from their surroundings, thus our method achieves remarkable results, outperforming the self-supervised DINO transformer backbone by a large margin. Visual examples are provided in \cref{fig:sup_quali_lunar}-right.
On the more challenging Lunar dataset where gray rocks are located on gray ground, the performance of our method still outperforms the one of DINO, see \cref{table:lunar}. We observe that our method tends to also segment shadows caused by uneven ground. Visual examples are provided in \cref{fig:sup_quali_lunar}-left.

\begin{figure}
\centering

{\footnotesize
\newcommand{\svgwidth}{0.72\linewidth}
%% Creator: Inkscape inkscape 0.92.5, www.inkscape.org
%% Accompanies image file '12210ad7-83f8-4b54-bb4b-e93f8ff6ac1f_q_horiz.pdf' (pdf, eps, ps)
\begingroup%
  \makeatletter%
  \providecommand\color[2][]{%
    \errmessage{(Inkscape) Color is used for the text in Inkscape, but the package 'color.sty' is not loaded}%
    \renewcommand\color[2][]{}%
  }%
  \providecommand\transparent[1]{%
    \errmessage{(Inkscape) Transparency is used (non-zero) for the text in Inkscape, but the package 'transparent.sty' is not loaded}%
    \renewcommand\transparent[1]{}%
  }%
  \providecommand\rotatebox[2]{#2}%
  \newcommand*\fsize{\dimexpr\f@size pt\relax}%
  \newcommand*\lineheight[1]{\fontsize{\fsize}{#1\fsize}\selectfont}%
  \ifx\svgwidth\undefined%
    \setlength{\unitlength}{1457.99993079bp}%
    \ifx\svgscale\undefined%
      \relax%
    \else%
      \setlength{\unitlength}{\unitlength * \real{\svgscale}}%
    \fi%
  \else%
    \setlength{\unitlength}{\svgwidth}%
  \fi%
  \global\let\svgwidth\undefined%
  \global\let\svgscale\undefined%
  \makeatother%
  \begin{picture}(1,0.38959421)%
    \lineheight{1}%
    \setlength\tabcolsep{0pt}%
    \put(0,0){\includegraphics[width=\unitlength,page=1]{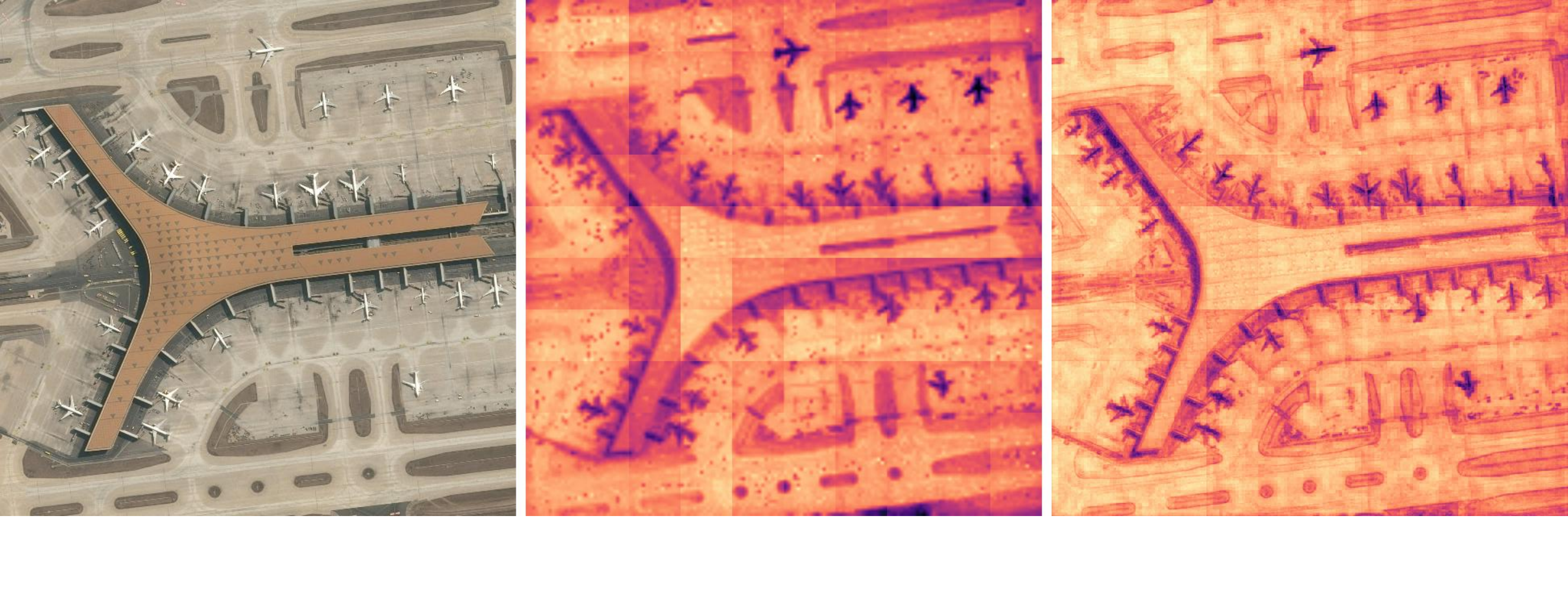}}%
    \put(0.50228982,0.01871055){\color[rgb]{0,0,0}\makebox(0,0)[t]{\lineheight{1.25}\smash{\begin{tabular}[t]{c}Averaged entropy\\ of SETR\end{tabular}}}}%
    \put(0.83288165,0.01997255){\color[rgb]{0,0,0}\makebox(0,0)[t]{\lineheight{1.25}\smash{\begin{tabular}[t]{c}Averaged entropy\\ of Segformer\end{tabular}}}}%
    \put(0.16518617,0.01131686){\color[rgb]{0,0,0}\makebox(0,0)[t]{\lineheight{1.25}\smash{\begin{tabular}[t]{c}Input\end{tabular}}}}%
    \put(0,0){\includegraphics[width=\unitlength,page=2]{12210ad7-83f8-4b54-bb4b-e93f8ff6ac1f_q_horiz.pdf}}%
  \end{picture}%
\endgroup%

}%
\\
% \includegraphics[width=0.45\linewidth]{images/quali_aircraft/12210ad7-83f8-4b54-bb4b-e93f8ff6ac1f_q_horiz.jpg}
% \hspace{0.3cm}
\vspace{10pt}
{\footnotesize
\newcommand{\svgwidth}{0.72\linewidth}
%% Creator: Inkscape inkscape 0.92.5, www.inkscape.org
%% Accompanies image file '22291e0b-ebe2-4f3f-b53e-4e709179300a_q_horiz.pdf' (pdf, eps, ps)
\begingroup%
  \makeatletter%
  \providecommand\color[2][]{%
    \errmessage{(Inkscape) Color is used for the text in Inkscape, but the package 'color.sty' is not loaded}%
    \renewcommand\color[2][]{}%
  }%
  \providecommand\transparent[1]{%
    \errmessage{(Inkscape) Transparency is used (non-zero) for the text in Inkscape, but the package 'transparent.sty' is not loaded}%
    \renewcommand\transparent[1]{}%
  }%
  \providecommand\rotatebox[2]{#2}%
  \newcommand*\fsize{\dimexpr\f@size pt\relax}%
  \newcommand*\lineheight[1]{\fontsize{\fsize}{#1\fsize}\selectfont}%
  \ifx\svgwidth\undefined%
    \setlength{\unitlength}{1457.99993079bp}%
    \ifx\svgscale\undefined%
      \relax%
    \else%
      \setlength{\unitlength}{\unitlength * \real{\svgscale}}%
    \fi%
  \else%
    \setlength{\unitlength}{\svgwidth}%
  \fi%
  \global\let\svgwidth\undefined%
  \global\let\svgscale\undefined%
  \makeatother%
  \begin{picture}(1,0.38959421)%
    \lineheight{1}%
    \setlength\tabcolsep{0pt}%
    \put(0,0){\includegraphics[width=\unitlength,page=1]{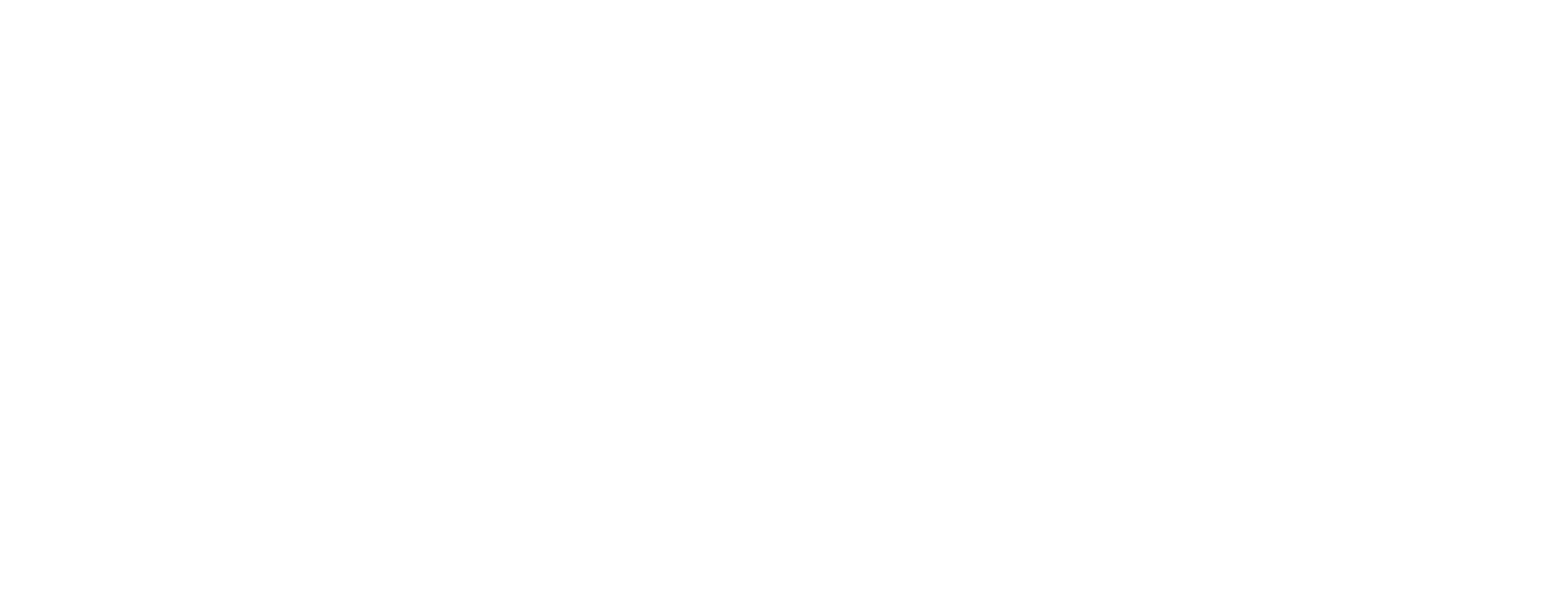}}%
    \put(0.50228982,0.01871055){\color[rgb]{0,0,0}\makebox(0,0)[t]{\lineheight{1.25}\smash{\begin{tabular}[t]{c}Averaged entropy\\ of SETR\end{tabular}}}}%
    \put(0.83288165,0.01997255){\color[rgb]{0,0,0}\makebox(0,0)[t]{\lineheight{1.25}\smash{\begin{tabular}[t]{c}Averaged entropy\\ of Segformer\end{tabular}}}}%
    \put(0.16518617,0.01131686){\color[rgb]{0,0,0}\makebox(0,0)[t]{\lineheight{1.25}\smash{\begin{tabular}[t]{c}Input\end{tabular}}}}%
    \put(0,0){\includegraphics[width=\unitlength,page=2]{22291e0b-ebe2-4f3f-b53e-4e709179300a_q_horiz.pdf}}%
  \end{picture}%
\endgroup%

}
\caption{
Qualitative results on Aircraft~\cite{AirbusAircraftSegmentation} dataset.
}
\label{fig:quali_aircraft}
\end{figure}

Furthermore, we also provide visual examples for the Aircraft dataset~\cite{AirbusAircraftSegmentation}, where no segmentation ground truth is available, see \cref{fig:quali_aircraft}. In this setting with a completely different viewing angle, our method still segments the objects, as the planes clearly stand out in the entropy maps. We expect this behavior to extend to most small salient objects that are visually distinct from the surrounding environment.

\subsection{Spatial Attention Study} \label{sec:exp_layers}
{
The choice of training set impacts the usefulness of attention entropy for object segmentation.
Extra semantic segmentation training improves the attention-entropy compared to ImageNet pre-training, presumably by making the transformer more sensitive to spatial information. However, self-supervised DINO’s class-token attention of the last layer can segment foreground objects, but its intermediate layer attention does not exhibit the clear object segmenting property we observe in semantic segmentation networks. Qualitative and quantitative comparisons are given in the supplementary material. Furthermore, we study the importance of the attention of individual layers and observe that intermediate layers' attentions as well as the average of all layers' attentions provide useful information for object segmentation. For practical purposes layers can be selected automatically  without any information about the test datasets, as presented in the supplementary material as well.

}

\section{Conclusion}\label{sec:conclusion}
We performed an in-depth study of the spatial attentions of vision transformers trained in a supervised manner for semantic segmentation \cite{\citeSETR,\citeSegformer,\citeViT}. For image patches associated to salient objects, the attention of intermediate layers tends to concentrate on that object, which we quantified via the Shannon entropy. On the other hand, for larger areas of coherent appearance, the entropy diffuses around the given image patch under consideration. This observation holds for many different layers of the transformer backbone and averaging their attentions improves the segmentation performance. We demonstrated that the attention in self-supervised trained transformers such as DINO exhibits a different behavior. Our observations hold for different degrees of domain shift, leading to a method that can segment unknown objects of small to moderate size in unknown context. We demonstrated this on several datasets with varying degrees of domain shift, up to a complete change of domain.  
Due to the negligible computational overhead with respect to the supervised model, our method might be suitable for automated driving applications, robotics applications and also for pre-segmenting objects for the sake of data annotation.

\section*{Acknowledgements}
S. Honari was involved in the present work during his employment at EPFL Lausanne.
A. Mütze and M. Rottmann acknowledge support by the German Federal Ministry of Education and Research within the junior research group project “UnrEAL” (grant no. 01IS22069).

\printbibliography

\appendix
\subsection{Transformer block}\label{sup:transdformerblock}
{
In addition to the description of the attention matrices in the paper, we describe the complete transformer block of the ViT~\cite{\citeViT} in the following. 
Let  $Z^{l-1}\in \mathbb{R}^{N^2 \times C}$, where $C$ denotes the number of feature channels of the feature map in layer $l\in \mathbb{N}$ and $N$ the number of image patches. Query, Key and Value are given by 
\begin{equation}
	\mathit{Q} = Z^{l-1} W_Q, ~~
	\mathit{K} = Z^{l-1} W_K, ~~ 
	\mathit{V} = Z^{l-1} W_V,
\end{equation} with learnable weight matrices $W_Q, W_K, W_V \in \mathbb{R}^{C \times d}$, $d = C/m$ and $m$ the number of attention heads.
Based on the attention matrices $A^l(Z^{l-1}) = \mathrm{softmax}( Q \cdot K^T / \sqrt{d} )$, we derive the corresponding self-attention $\mathit{SA}$. It is expressed as}
\begin{equation}
	\mathit{SA}^l (Z^{l-1}) = A^l (Z^{l-1}) \cdot \mathit{V} \, .
\end{equation}
Again, accounting for the multiple heads, this results in $m$ independent self-attention matrices $\mathit{SA}^l_i$, $i=1,\ldots,m$, that are concatenated 
\begin{equation}
    \mathit{SA}^l_{1:m}(Z^{l-1}) = [\mathit{SA}^l_1(Z^{l-1}),\ldots,\mathit{SA}^l_m(Z^{l-1})]   
\end{equation}
and then multiplied with another weight matrix $W_O  \in \mathbb{R}^{md \times C}$.
Combined with a skip connection, this gives
\begin{equation}
 \mathit{MSA}^l(Z^{l-1}) =  \mathit{SA}^l_{1:m}(Z^{l-1}) + \mathit{SA}^l_{1:m}(Z^{l-1}) W_O \, .
\end{equation}
Finally, $\mathit{MSA}^l(Z^{l-1})$ is again input to a multilayer perceptron ($\mathit{MLP}$) block with a skip connection. This yields
\begin{equation}
  Z^{l} = \mathit{MSA}^l(Z^{l-1}) + \mathit{MLP}(\mathit{MSA}^l(Z^{l-1})) \in \mathbb{R}^{N^2 \times C} \, .
\end{equation}
{
The implementation of the attention extraction is given in \cref{sup:AttentionExtraction}.
}

\subsection{Attention Entropy Extraction}\label{sup:AttentionExtraction}

In our experiments, we evaluate the semantic segmentation transformers SETR~\cite{\citeSETR} and Segformer~\cite{\citeSegformer} implemented in the MMSegmentation framework~\cite{mmseg2020}. Both have been trained for semantic segmentation with the Cityscapes~\cite{Cordts16} dataset. As both architectures are designed to operate on square images, sliding window inference is automatically performed on our rectangular images.

In the case of SETR~\cite{\citeSETR}, we extract the attention maps from its ViT backbone. We discard the class token leaving the patch-to-patch attention and calculate the Shannon entropy for each patch.
The entropy from each layer has a resolution of $48 \times 48$, and we perform linear interpolation to upscale it to the input image resolution.
We use the {\it PUP} variant of the SETR decoder head as it has the best reported segmentation performance.
The DPT~\cite{Ranftl21dpt} network has a ViT backbone like SETR and we perform the extraction in the same manner.

The Segformer~\cite{\citeSegformer} uses a pyramid of self-attention operations with the base of the pyramid comprising $256^2$ patches.
We use the {\it MIT-B3} variant as a compromise between performance and computation cost.  It would not be feasible to calculate attention between each pair of patches. Hence it performs a linear operation reducing $r$-times the number of rows of $K$ and $V$, where $r$ is a chosen reduction ratio. As a consequence, the attention matrices $A^l_i$ have only $\frac{N^2}{r}$ rows instead of $N^2$. This essentially means the attention map's resolution is reduced. Nevertheless, we can calculate entropy over each row of $\overline{A}^l$, upscale the heatmaps to a common shape and average them. The number of tokens, and thus the resolution of the entropy maps, varies across layers from $256 \times 256$ to $32 \times 32$. We linearly interpolate all of them to $256 \times 256$ before averaging, then resize the average heatmap to full image resolution.

\subsubsection{Hardware and versions.}

The experiments were done on a V100 GPU.
The software used was PyTorch 2.0.1 and the MMSegmentation framework~\cite{mmseg2020} version 1.1.1.

\subsubsection{Compuational Cost.}

The calculation of attention-entropy causes only a small overhead over the backbone and semantic-seg model:
\begin{itemize}
\item $(1.66 \pm 0.04) \%$ for SETR,
\item $(3.76 \pm  0.66) \%$ for DPT,
\item $(7.3 \pm 0.5) \%$ for Segformer.
\end{itemize}

\subsection{Interactive attention visualization}\label{sup:InteractiveVis}

To view the interactive attention visualization, please
open \url{https://liskr.net/attentropy/} with a web browser.
Hover the cursor over the input image to choose the source patch $j$.
The heatmap will display the attention values from the chosen patch $j$ to each other patch $j'$, that is the value $\bar{A}^{l}_{j,j'}$.
The attention values are extracted from the ViT~\cite{\citeViT} backbone of the SETR~\cite{\citeSETR} network,
and have been truncated to the $[0, 0.005]$ range for visual readability.
\begin{itemize}
\item By selecting a source patch within an object, we can observe how its attention is concentrated.
In contrast, the road and background areas exhibit diffuse attention.
This coincides with lower entropy of the attention map.
\item Choose the layer shown using left and right arrow keys.
We can observe how the earlier layers' attention is spatially localized,
and therefore useful for small object segmentation.
The later layers show less localized attention and so we exclude them from the entropy average.
\end{itemize}

\noindent 
Note, that we compressed the images due to storage limitations.

\subsection{Attention layers}

\subsubsection{Layer subset selection}
\begin{figure} 
	\centering
        \parbox{0.39\linewidth}{
        \centering
		\includegraphics[width=0.55\linewidth]{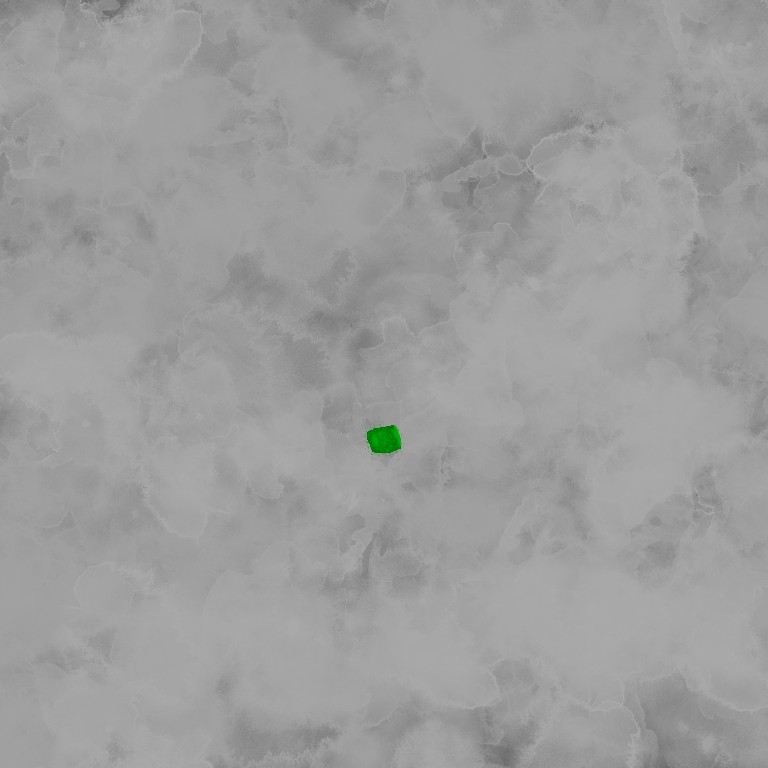} } 
        \hfill
        \parbox{0.55\linewidth}{
        \centering
        \includegraphics[width=0.70\linewidth]{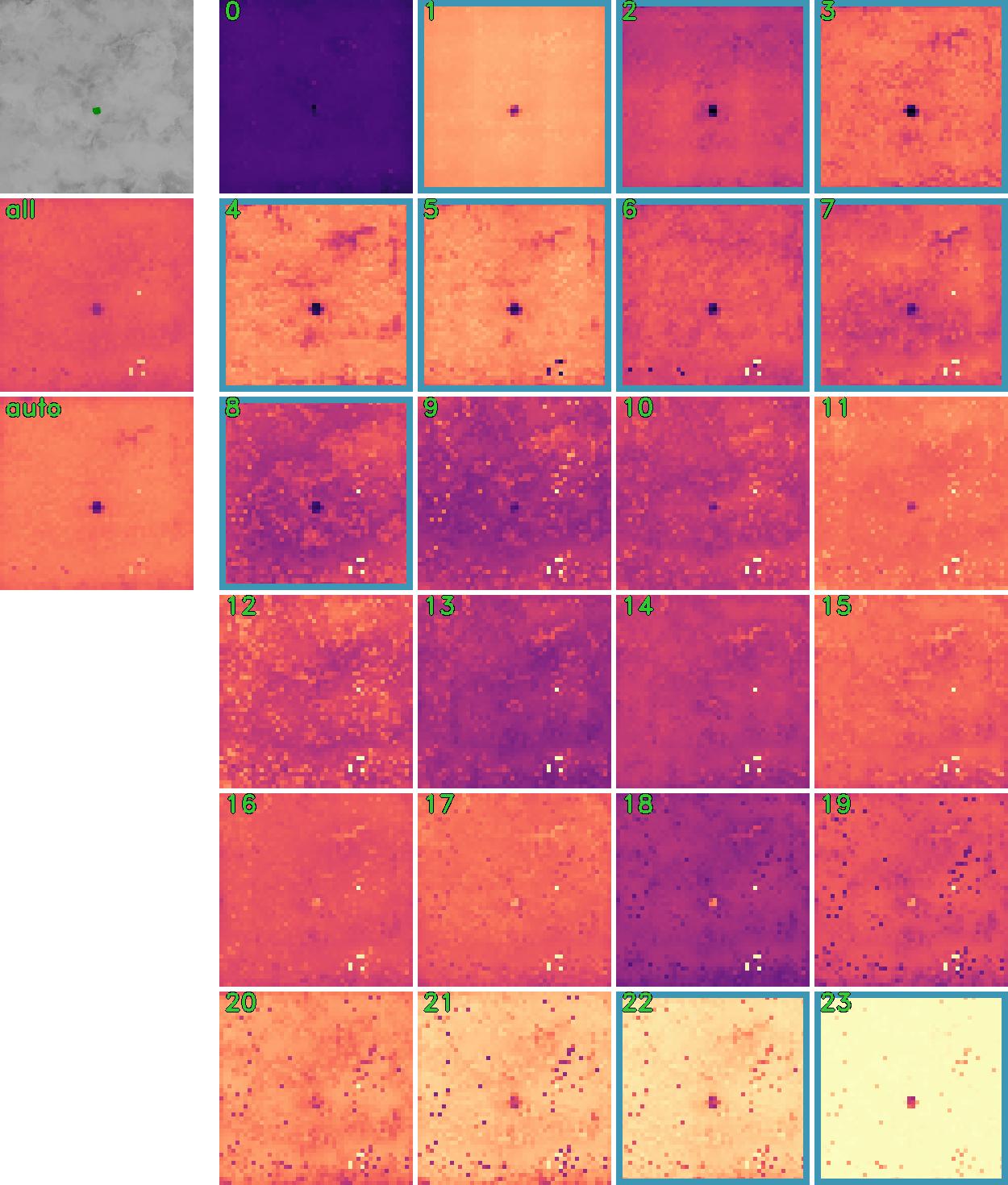}}\\	
        \parbox{0.41\linewidth}{(a) Test pattern simulating a small object, used to select a subset of layers whose attention-entropy is most useful.}  
        \hfill
        \parbox{0.55\linewidth}{(b) Attention entropy of each layer of SETR given the test pattern above as input.}\\
        
        \vspace{5pt}
        {\includegraphics[width=0.75\linewidth]{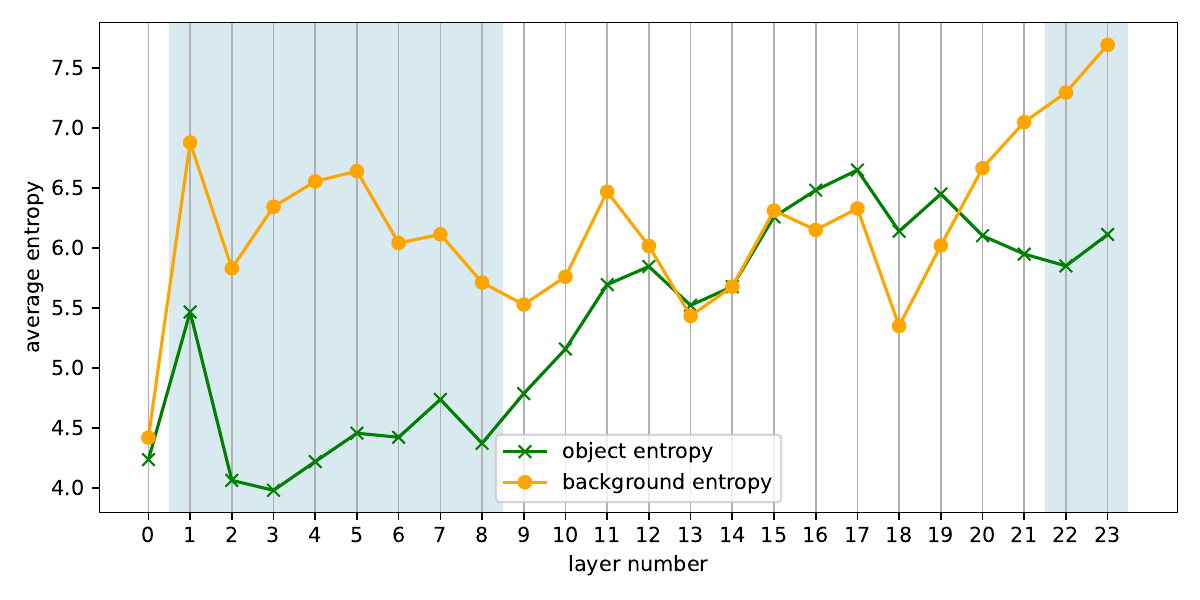}} \\
        \parbox{\linewidth}{(c) Comparison of average entropy within the object pixels and the background pixels. 
        The chosen layers (where background entropy is 20\% higher than object entropy) are highlighted in blue.}
		\vspace{1pt}

	% \begin{subfigure}[b]{\linewidth}
	% 	\centering
	% 	\includegraphics[width=\linewidth]{images/auto_layer_selection/TexTest5_VitLarge-SETR-ctc_f1.2__layers.jpg}
	% 	\caption{
	% 	Attention entropy of each layer of SETR given the test pattern above as input.
	% 	}
	% 	\label{fig:layers_setr_testpattern}
	% \end{subfigure}
	% \begin{subfigure}[b]{\linewidth}
	% 	\centering
	% 	\includegraphics[width=\linewidth]{images/auto_layer_selection/TexTest5_VitLarge-SETR-ctc_f1.2__levels.pdf}
	% 	\caption{
	% 		Comparison of average entropy within the object pixels and the background pixels.
	% 		The chosen layers (where background entropy is 20\% higher than object entropy) are highlighted in blue.
	% 	}
	% 	\label{fig:layers_setr_plot}
	% \end{subfigure}
	%
	%% \includegraphics[width=0.8\linewidth]{images/auto_layer_selection/TexTest5_VitLarge-SETR-ctc_f1.2__layers.jpg}
	%% \includegraphics[width=0.98\linewidth]{images/auto_layer_selection/TexTest5_VitLarge-SETR-ctc_f1.2__levels.pdf}	
	%% \includegraphics[width=0.48\linewidth]{images/auto_layer_selection/TexTest5_VitBase-DPT-ade20k_f1.2__layers.jpg}
	%% \includegraphics[width=0.5\linewidth]{images/auto_layer_selection/TexTest5_VitLarge-SETR-ctc_f1.2__layers.jpg}
	\caption{
	Procedure for selecting a subset of layers whose attention-entropy is most useful.
	This example uses the SETR backbone.
	}
	\label{fig:auto_layers} \label{fig:testpattern} \label{fig:layers_setr_testpattern} \label{fig:layers_setr_plot}
\end{figure}

\begin{figure}
    \begin{tabular}{cc}
	\centering
	% \begin{subfigure}[b]{\linewidth}   	
        \centering
		\includegraphics[width=0.45\linewidth]{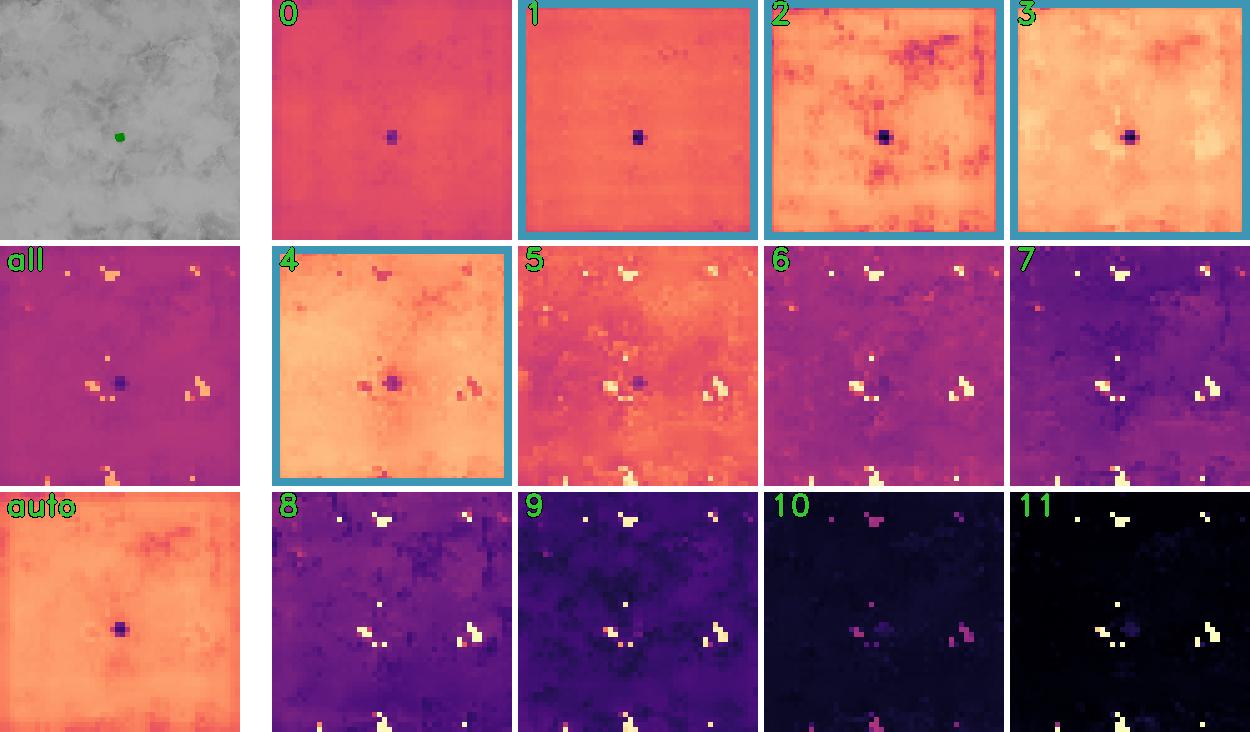} 
        &
        \includegraphics[width=0.49\linewidth]{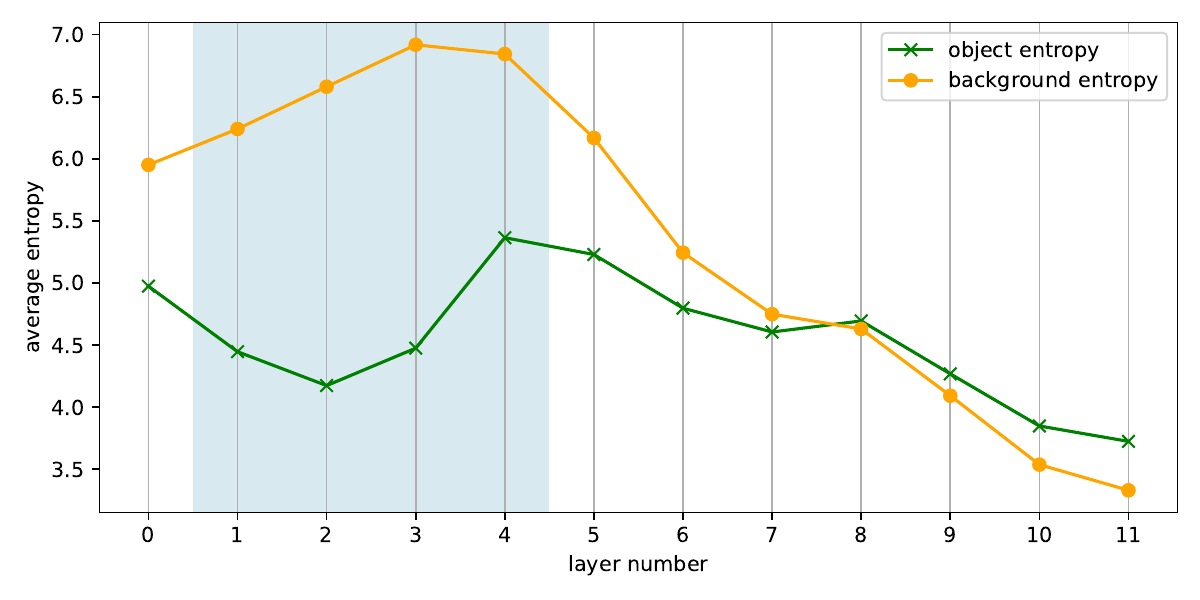} \\
		
		\parbox{0.45\linewidth}{(a) Attention entropy of each layer of DPT given the test pattern as input.} &\parbox{0.45\linewidth}{ (b) Entropy layers within object and background.}
	\end{tabular}
		%% \label{fig:layers_setr_testpattern}
	% \begin{figure}
	% 	\centering
	% 	\includegraphics[width=\linewidth]{images/auto_layer_selection/TexTest5_VitBase-DPT-ade20k_f1.2__levels.pdf}
	% 	\caption{
	% 		Entropy layers within object and background.
	% 	}
	% 	% \label{fig:layers_setr_plot}
	% \end{subfigure}
	% %
	%% \includegraphics[width=0.8\linewidth]{images/auto_layer_selection/TexTest5_VitLarge-SETR-ctc_f1.2__layers.jpg}
	%% \includegraphics[width=0.98\linewidth]{images/auto_layer_selection/TexTest5_VitLarge-SETR-ctc_f1.2__levels.pdf}	
	%% \includegraphics[width=0.48\linewidth]{images/auto_layer_selection/TexTest5_VitBase-DPT-ade20k_f1.2__layers.jpg}
	%% \includegraphics[width=0.5\linewidth]{images/auto_layer_selection/TexTest5_VitLarge-SETR-ctc_f1.2__layers.jpg}
	\vspace{1pt}
    \caption{
	Procedure of layer selection for the DPT network.
	}
	\label{fig:auto_layers_dpt}
 
\end{figure}

Transformer backbones contain multiple layers of self-attention.
Not all layers have equally useful attention entropy patterns for small object segmentation.
We have observed that averaging only a subset of layers can improve the performance of the entropy-based segmentation.
We choose the layers in a way that does not use any information about the test datasets and objects.
To that end we create a purely synthetic test pattern with simple textures,
shown in \cref{fig:testpattern}-(a), drawn using an image editing program. 
We then calculate the attention entropy of each layer given the test pattern as input, as shown in \cref{fig:layers_setr_testpattern}-(b).
We measure the average entropy within the object pixels and compare it to the entropy averaged over the background region.
The levels are illustrated in \cref{fig:layers_setr_plot}-(c).
We choose the subset of layers where background entropy is 20\% higher than object entropy.
This way lower entropy consistently occurs in objects.
Averaging the entropy values creates a useful signal for delineating small objects.
The procedure is automatic and can be performed for any new model, in \cref{fig:auto_layers_dpt} we show the results for DPT.

\subsubsection{Layer-Wise Study.} 
\begin{table*}[t]
\centering
\setlength{\heavyrulewidth}{1pt}
\resizebox{0.85\textwidth}{!}{
\begin{tabular}{ll|rr|rr|rr|rr}
	\toprule
{} & {} & \multicolumn{2}{c|}{Lost and Found} & \multicolumn{2}{c|}{Road Obstacles 21} & \multicolumn{2}{c|}{MaSTr1325} & \multicolumn{2}{c}{Artificial Lunar} \\
{} & {} & \multicolumn{2}{c|}{test no known} & \multicolumn{2}{c|}{test} & \multicolumn{2}{c|}{} & \multicolumn{2}{c}{Landscape} \\
{} & {} & \segmetricsShort & \segmetricsShort & \segmetricsShort & \segmetricsShort \\
	\midrule
\multirow{2}{*}{\shortstack{SETR}}
{} & subset & \textbf{73.0} & \textbf{3.9} & \textbf{71.3} & \textbf{2.5} & 59.6 & 39.6 & 34.7 & 87.6 \\
{} & all &	62.2 & 11.2 & 67.0 & 5.8 & 32.3 & 54.0 & - & - \\
\midrule
\multirow{2}{*}{\shortstack{Segformer}}
{} & subset & 51.2 & 10.8 & 32.1 & 11.5 & 55.8 & 51.0 & 34.0 & \textbf{87.5} \\
{} & all & 47.4 & 15.6 & 34.7 & 12.0 & - & - & - & - \\
\midrule
\multirow{2}{*}{\shortstack{DPT}}
{} & subset & 57.0 & 5.7 & 55.1 & 3.8 & {\bf 71.9} & {\bf 14.2} & {\bf 35.1} & 89.9 \\
{} & all & 32.7 & 28.0 & 42.6 & 44.5 & 30.0 & 80.0 & 21.5 & 92.2 \\
\bottomrule
\end{tabular}
}	
\caption{Obstacle detection scores for layer averaging strategies.} 
% The primary metrics are Average Precision (AP$\uparrow$) for pixel classification 
% and Average $F_1$ ($\overline{F_1}\uparrow$) for instance level detection.
% The evaluation protocol and metrics follow the {\it Segment Me If You Can} benchmark~\cite{Chan21b}.
% \KL{SETR with regression may be worse than SETR manual for ObstacleTrack because the regression takes an average loss over two datasets, so it is a different compromise.}
\label{table:benchmarks_ablation}
\end{table*}

\begin{figure} 
\centering
\includegraphics[width=0.49\linewidth]
{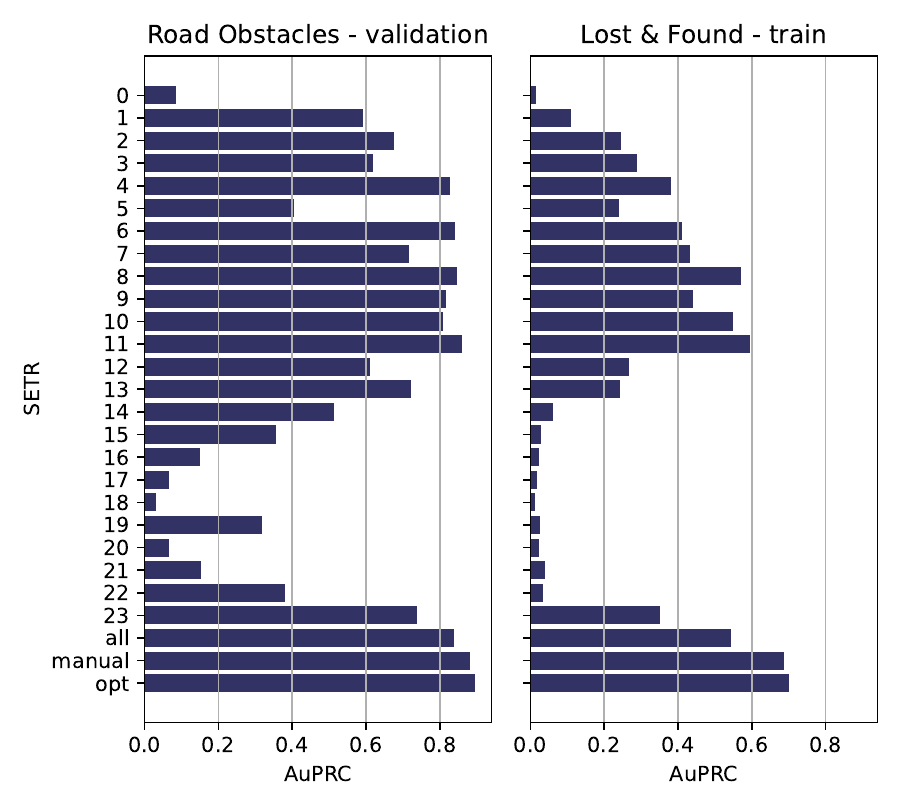}
\includegraphics[width=0.49\linewidth]
{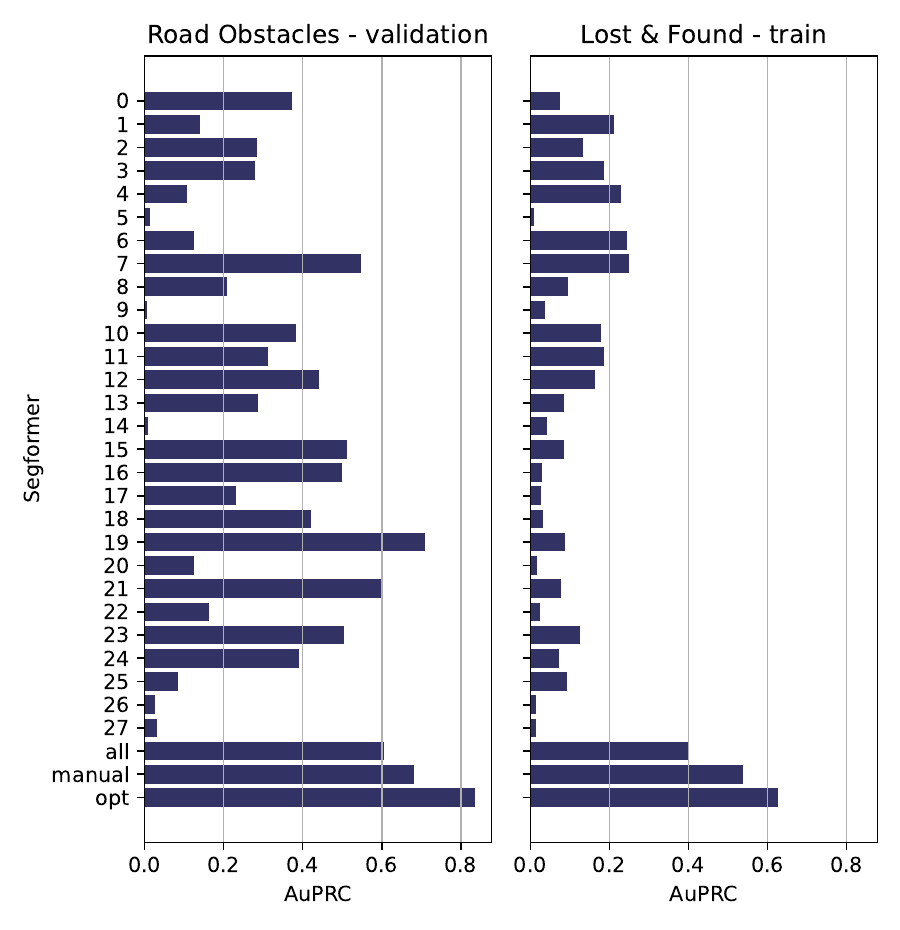}
\caption{Usefulness of individual layers' entropy for obstacle detection,
measured by the area under the precision-recall curve of obstacle segmentation on {\it Road Obstacles 21 - validation} and {\it Lost and Found - train} datasets.
% In addition to layers denoted by their 0-based index, we list the scores for layer averaging strategies discussed in Sec.~\ref{sec:exp_layers}.
% {\it all}: average of all layers; {\it manual}: average of a manually selected subset of layers;  
% {\it optimized}: weighted average determined through logistic regression.
% We observe that averaging overall improves obstacle detection quality over individual layers, particularly so for the Segformer.
% \KL{TODO Highlight manually selected layers in a different color?}
}
\label{fig:layer_scores}
\end{figure}

We study the usefulness of individual layers' entropy heatmaps for obstacle detection.
We measure the area under the precision-recall curve of obstacle segmentation on the {\it Road Obstacles} {\it validation} set and the {\it Lost and Found} training set -- those datasets have publicly available obstacle labels and do not intersect with the benchmark test sets used for  comparisons. 
We plot the layers' scores in Fig.~\ref{fig:layer_scores}, together with scores calculated for the following averaging strategies:
\begin{itemize}
	\item{\it all layers}: the output heatmap is the average of all attention-entropy heatmaps $\mathbf{E}^l$, that is $\mathcal{L} = \{1,\ldots,l\}$.
	\item{\it manual layers}: we average a subset of layers whose entropy corresponds well to obstacles, as confirmed by visual inspection of example outputs. The chosen layers are shown in the supplementary material.
    %highlighted in Fig.~\ref{fig:entropy_layers}.
	\item{\it optimized weights}: we use logistic regression to determine the optimal weights to mix the layer entropy heatmaps, that is $\sigma(\sum_{l \in \mathcal{L}} a_l \mathbf{E}^l + b \mathbf{1})$, where $\sigma$ is the sigmoid activation, $a_l,b \in \mathbb{R}$ and $\mathbf{1} \in \mathbb{R}^{N \times N}$ is a matrix with all elements equal to one.
	The weights are obtained using the frames from {\it Road Obstacles - Validation} and {\it Lost and Found - train}.
\end{itemize}
The results show that while averaging entropy across all layers obtains decent results, a simple visual choice of layers yields further improvement in manual layers. The regression weighting obtains either similar results as manual (for SETR) or further improves compared to it (for Segformer). We used the manual selection in our comparisons as it does not apply any fine-tuning or learning on top of pre-trained network. We additionally compare the layer averaging strategies in Tab.~\ref{table:benchmarks_ablation} for DPT and the distant domains.

\begin{figure} 
    \centering
    \includegraphics[height=0.48\linewidth]{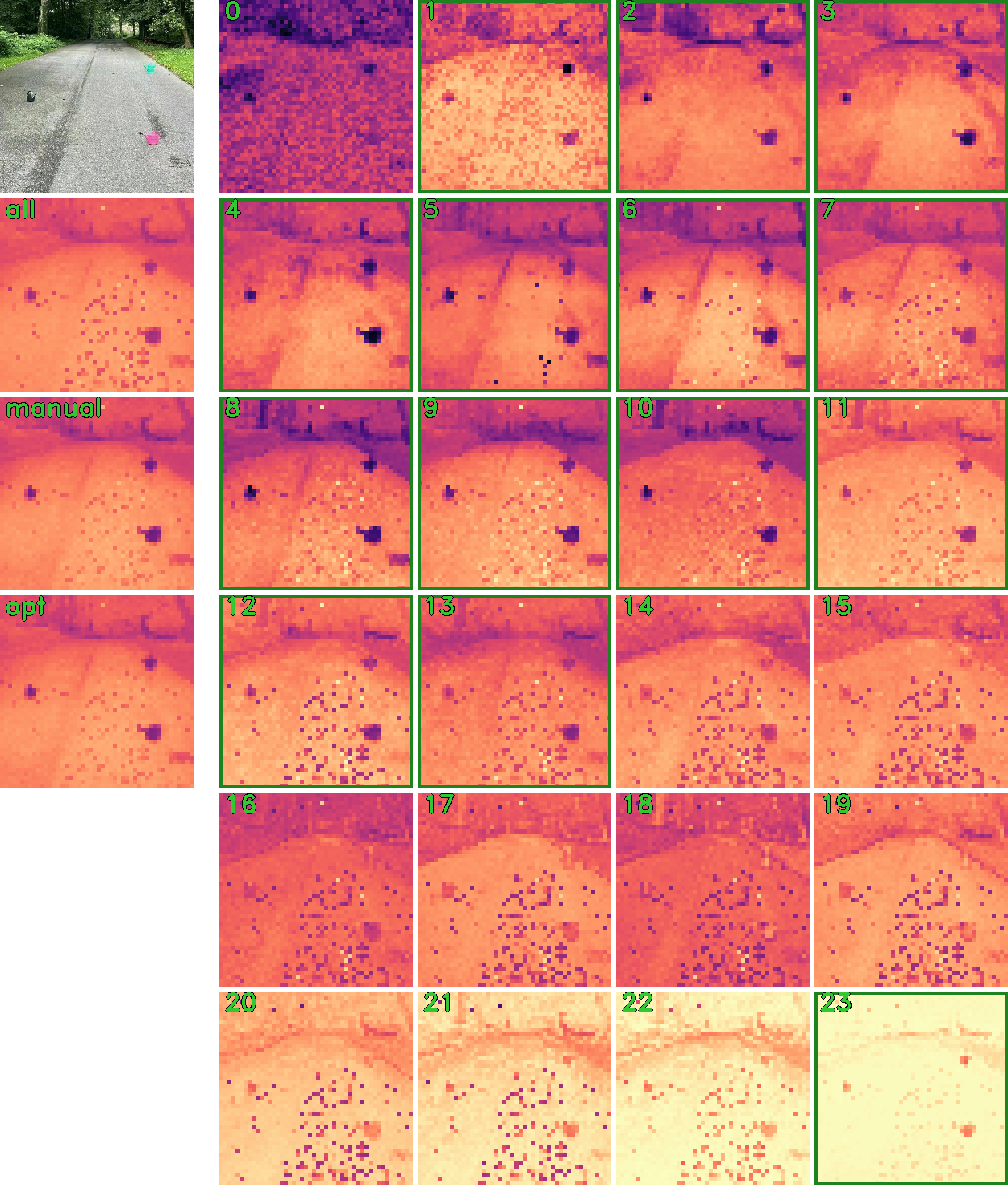}
    \hspace{10mm}
    \includegraphics[height=0.48\linewidth]{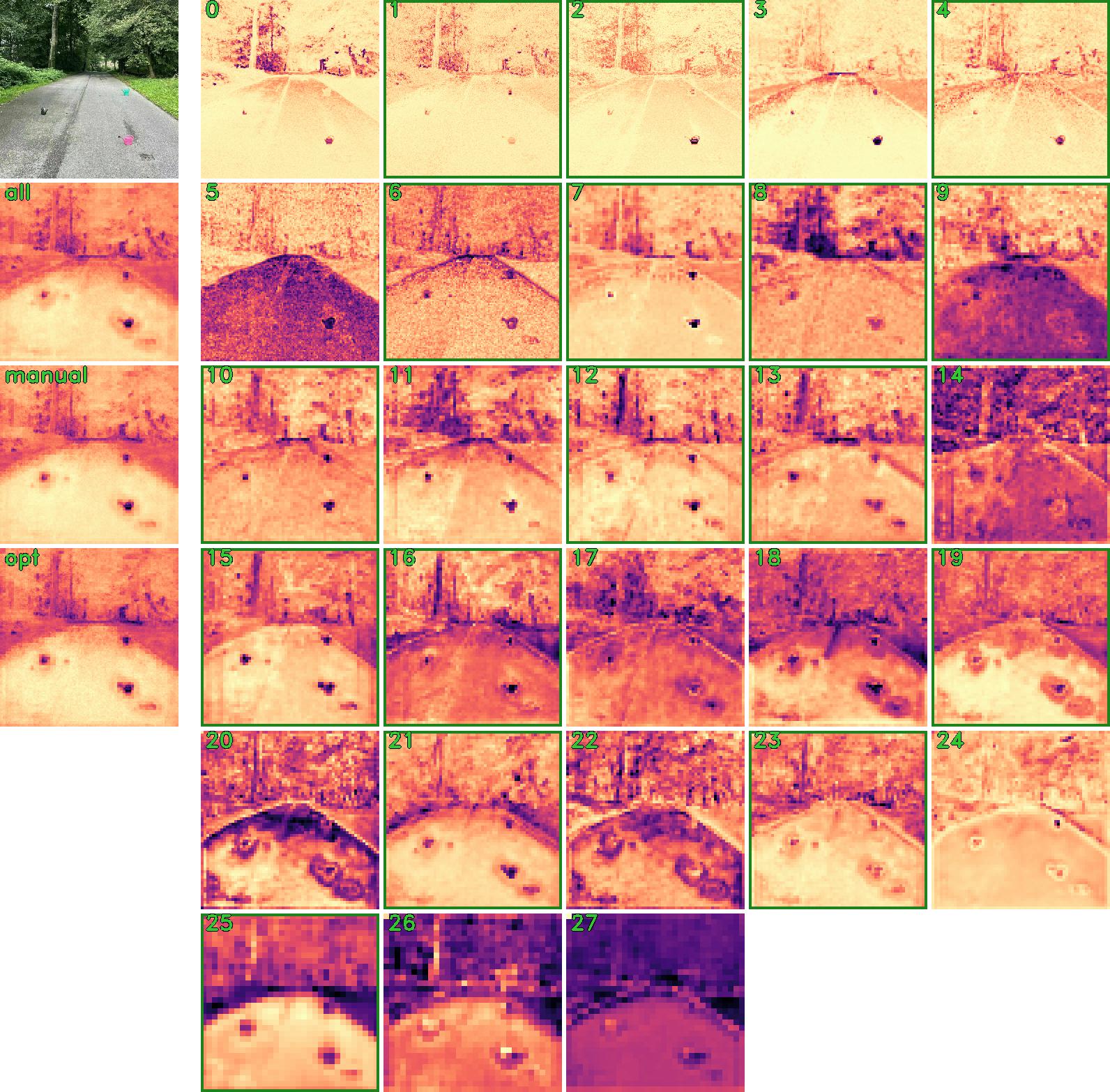}
    \\
    \makebox[0.475\linewidth]{\small SETR (Visual Transformer)} \makebox[0.475\linewidth]{\small Segformer}
    \caption{
    Attention-entropy heatmaps for layers of SETR's Visual Transformer and Segformer backbone
    for a real image from the Road Obstacle dataset.
    }
    \label{fig:entropy_layers}
\end{figure}

\subsubsection{Training impact on Attention}
\begin{figure} 
        \centering
        \includegraphics[width=0.24\linewidth]{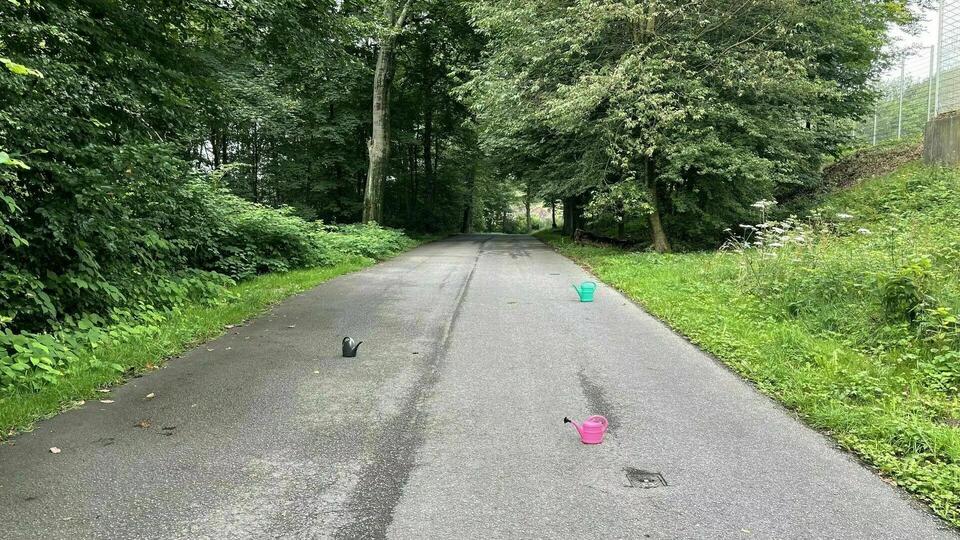}
        \includegraphics[width=0.24\linewidth]{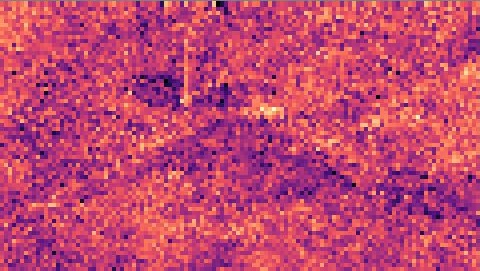}        \includegraphics[width=0.24\linewidth]{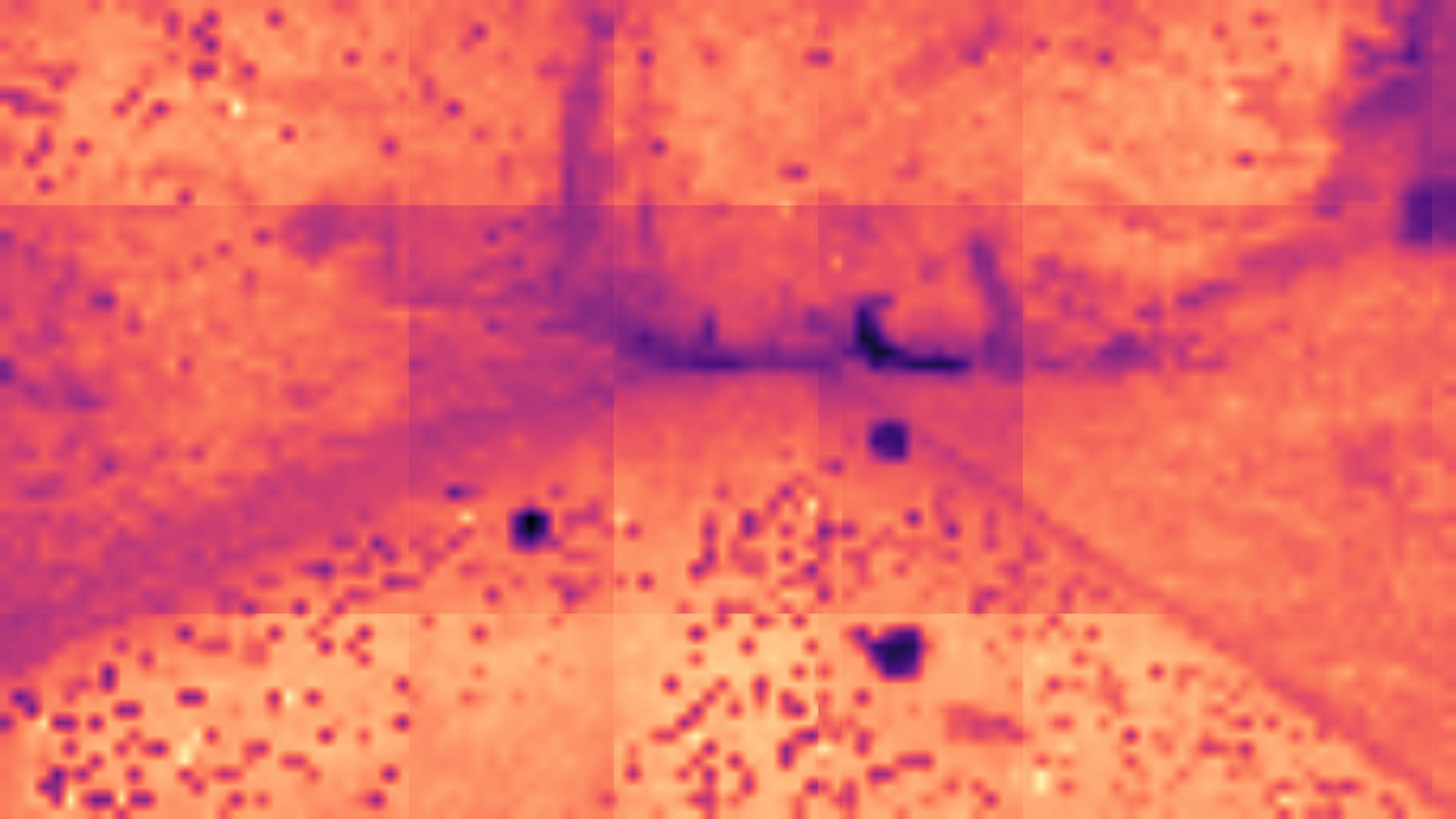}
        \includegraphics[width=0.24\linewidth]{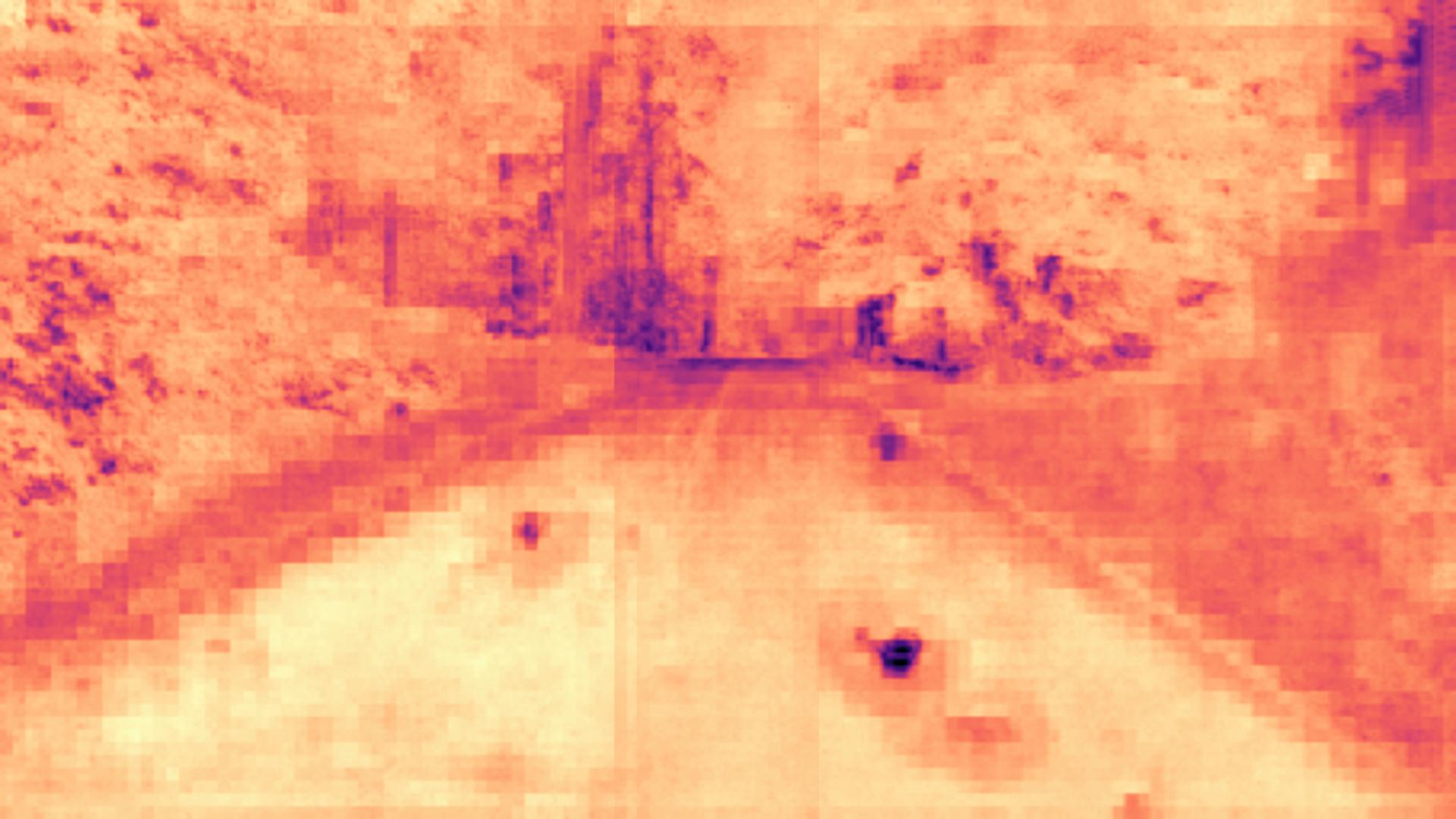}\\
        \makebox[0.24\linewidth]{\scriptsize Input} \makebox[0.24\linewidth]{\scriptsize DINO~\cite{caron2021emerging} entropy sum} 
        % \vspace{5pt}
        \makebox[0.24\linewidth]{\scriptsize SETR entropy sum}\makebox[0.24\linewidth]{\scriptsize Segformer entropy sum}
        \caption{
        Sums of attention-entropy over all layers - comparison between
        DINO~\cite{caron2021emerging} trained in a self-supervised manner on ImageNet
        versus SETR~\cite{\citeSETR} and Segformer~\cite{\citeSegformer} trained for Cityscapes semantic segmentation.
        }
        \label{fig:dino_attentropy}
    \end{figure}

The choice of training setup impacts the usefulness of attention entropy for object segmentation.
We compare ViT trained in four ways:
\begin{itemize}
	\item ImageNet supervised pre-training,
	\item ImageNet supervised pre-training followed by Cityscapes segmentation fine-tuning (SETR),
	\item ImageNet supervised pre-training followed by ADE20k segmentation fine-tuning (DPT),
	\item ImageNet unsupervised pre-training with DINO~\cite{caron2021emerging}.
\end{itemize}
The detection scores are given in Tab.~\ref{table:ablation_training}. 
{
The findings are that explicitly training a semantic segmentation task (first two rows) improves the attention-entropy compared to ImageNet pre-training (third row). It can be assumed that this was achieved by enhancing the transformer's sensitivity to spatial information. Comparing the self-supervised DINO~\cite{caron2021emerging} transformer (last row) and the transformers trained for Cityscapes segmentation (first row) the results show a notable performance gap for all datasets. Both DINO and SETR have a ViT~\cite{\citeViT} network as their backbones.
Despite the capacity of the self-supervised DINO to segment foreground objects through its class-token attention in the final layer, its intermediate layer attention does not display the distinctiveness in object segmentation observed in semantic segmentation networks. This is exemplified in \cref{fig:dino_attentropy}.
}
\begin{table*}[tb]
\centering
\setlength{\heavyrulewidth}{1pt}
\resizebox{\textwidth}{!}{
\scriptsize{
\begin{tabular}{l|rr|rr|rr|rr}
\toprule
{} & \multicolumn{2}{c|}{Lost and Found} & \multicolumn{2}{c|}{Road Obstacles 21} & \multicolumn{2}{c|}{MaSTr1325} & \multicolumn{2}{c}{Artificial Lunar} \\
{} & \multicolumn{2}{c|}{test no known} & \multicolumn{2}{c|}{test} & \multicolumn{2}{c|}{} & \multicolumn{2}{c}{Landscape} \\
{} & \segmetricsShort & \segmetricsShort & \segmetricsShort & \segmetricsShort \\
\midrule
ImageNet + Cityscapes (SETR) & \textbf{73.0} & \textbf{3.9} & \textbf{71.3} & \textbf{2.5} & 59.6 & 39.6 & 34.7 & \textbf{87.6} \\
ImageNet + ADE20k (DPT) & 57.0 & 5.7 & 55.1 & 3.8 & {\bf 71.9} & {\bf 14.2} & {\bf 35.1} & 89.9 \\
ImageNet & 20.1 & 52.6 & 40.2 & 19.9 & 16.5 & 82.2 & 24.2 & 92.2 \\
ImageNet unsupervised with DINO & 15.7 & 48.4 & 16.4 & 75.1 & 21.3 & 74.8 & 21.0 & 94.3 \\
\bottomrule
\end{tabular}
}	}
\caption{\textbf{Comparison of entropy based on ViT training}. 
}
\label{table:ablation_training}
\end{table*}

{
\subsection{Obstacle detection scores}
    In addition to the training-free methods we also evaluate the performance of our approach compared to methods which use some kind of specific training to detect obstacles or anomalies. In the upper section of Tab.~\ref{table:benchmarks_baselines_full} we compare training-free methods, including our AttEntropy method as shown in the main paper. The lower section of the table reports training-based methods. Training-based methods have the potential to achieve a higher level of obstacle detection performance (see results in bold). However, our training-free method which utilizes the entropy of spatial attentions, achieves comparable results to the training-based method SynBoost\cite{DiBlase21}, and outperforms training-based methods like Road Inpainting \cite{Lis20} on the \textit{Road Obstacle 21} track. For a visual comparison of our method and SynBoost on more distant domains see \cref{sup:qualitative_examp}.
}

    \begin{table*}[t]
    \centering
    \setlength{\heavyrulewidth}{1pt}
    \setlength{\tabcolsep}{4pt}
    {
    
    \resizebox{0.85\textwidth}{!}{
        \begin{tabular}{ll|rrrrr|rrrrrlll}
        \toprule{} & {} & \multicolumn{5}{c|}{Lost and Found - test no known} & \multicolumn{5}{c}{Road Obstacles 21 - test} \\
        {} & {} & \segmetricsA{c|} & \segmetricsA{c} \\
        {} & {} & \segmetricsB & \segmetricsB \\
        \midrule
        
        \multirow{14}{*}{\shortstack{No \\ training}}{} & Ensemble \cite{Lakshminarayanan17} & 2.9 & 82.0 & 6.7 & 7.6 & 2.7 & 1.1 & 77.2 & 8.6 & 4.7 & 1.3 \\
        {} & Embedding Density \cite{Blum19} & 61.7 & 10.4 & 37.8 & 35.2 & 27.5 & 0.8 & 46.4 & 35.6 & 2.9 & 2.3  \\
        {} & LOST \cite{LOST} &1.1 & 94.7 & 8.6 & 6.0 & 6.0 & 4.7 & 93.8 &17.0 & 8.3 &11.0 \\
        {} & MC Dropout \cite{Mukhoti18} & 36.8 & 35.5 & 17.4 & 34.7 & 13.0 & 4.9 & 50.3 & 5.5 & 5.8 & 1.0 \\
        {} & Max Softmax \cite{Hendrycks17b} & 30.1 & 33.2 & 14.2 & 62.2 & 10.3 & 15.7 & 16.6 & 19.7 & 15.9 & 6.3 \\
        {} & Mahalanobis\cite{Lee18a} & 55.0 & 12.9 & 33.8 & 31.7 & 22.1 & 20.9 & 13.1 & 13.5 & 21.8 & 4.7 \\
        {} & ODIN \cite{Liang18b} & 52.9 & 30.0 & 39.8 & 49.3 & 34.5 & 22.1 & 15.3 & 21.6 & 18.5 & 9.4 \\
        {} & DINO \cite{caron2021emerging} & 26.4 & 38.9 &11.7 &13.6 & 5.7 & 39.9 & 14.1 &26.8 &19.1 &12.4 \\
        {} & M2F-EAM \cite{grcic2023advantages}& - & - & - & - & - & {66.9} & 17.9 & - & - & -\\
        {} & RbA \cite{nayal2023rba}& - & - & - & - & - & {\bf 87.8} & 3.3 & 47.4 & 56.2 & {\bf 50.4}\\
        {} & RAOS (trainig free) \cite{fu2023evolving} & {\bf 81.1} & 3.2 & 53.5 & 46.2 & {\bf 50.6} & \underline{79.7} & 0.8 & 42.4 & 32.6 & 32.7\\
        {} & Ours Segformer & 51.2 & 10.8 & 37.8 & 35.6 & 28.9 & 32.1 & 11.5 & 24.3 & 30.1 & 18.6 \\
        {} & Ours DPT & 57.0 & 5.7 & 17.2 & 31.8 & 18.3 & 55.1 & 3.8 & 20.8 & 42.8 & 24.1 \\
        {} & Ours SETR & \underline{73.0} & 3.9 & 35.5 & 43.9 & \underline{37.5} & 71.3 & 2.5 & 35.7 & 46.3 & \underline{39.7} \\
        \midrule
        
        \multirow{8}{*}{\shortstack{Obstacle \\ or anomaly \\ training}}{} & Void Classifier \cite{Blum19} & 4.8 & 47.0 & 1.8 & 35.1 & 1.9 & 10.4 & 41.5 & 6.3 & 20.3 & 5.4 \\
        {} & JSRNet \cite{Vojir21} & 74.2 & 6.6 & 34.3 & 45.9 & 36.0 & 28.1 & 28.9 & 18.6 & 24.5 & 11.0 \\
        {} & Image Resynthesis \cite{Lis19} & 57.1 & 8.8 & 27.2 & 30.7 & 19.2 & 37.7 & 4.7 & 16.6 & 20.5 & 8.4 \\
        {} & Road Inpainting \cite{Lis20} & 82.9 & 35.7 & 49.2 & 60.7 & 52.3 & 54.1 & 47.1 & 57.6 & 39.5 & 36.0 \\
        {} & SynBoost \cite{DiBlase21} & 81.7 & 4.6 & 36.8 & 72.3 & 48.7 & 71.3 & 3.2 & 44.3 & 41.8 & 37.6 \\
        {} & Max Entropy \cite{Chan21a} & 77.9 & 9.7 & 45.9 & 63.1 & 49.9 & 85.1 & 0.8 & 47.9 & 62.6 & 48.5 \\
        {} & DenseHybrid \cite{grcic2022densehybrid} &  78.7 & 2.1 & 46.9 & 52.1 & 52.3 & 87.1 & 0.2 & 45.7 & 50.1 & 50.7 \\
        {} & M2F-EAM \cite{grcic2023advantages} &  - & - & - & - & - & \underline{92.9} & 0.5 & 65.9 & 76.5 & {\bf 75. 6}\\
         {} & RbA \cite{nayal2023rba} &  - & - & - & - & - &  91.8 & 0.5 & 58.4 & 58.8 & 60.9\\
         {} & $\text{CSL}_2$ \cite{zhang2024csl}& {\bf 89.8} & 0.5 & 55.0 & - & {\bf 62.4} & 87.1 & 0.7 & 44.7 & - & 51.0\\
        {} & NFlowJS \cite{grcic2024dense} & \underline{89.3} & 0.65 & 54.6 & 59.7 & \underline{61.8} &  85.6 & 0.4 & 45.53 & 49.5 & 50.4\\
        {} & Mask2Anomaly \cite{rai2023unmasking} &  - & - & - & - & - &  {\bf 93.2} & 0.2 & 55.7 & 75.4 & \underline{68.2}\\
        {} & DaCUP \cite{vojivr2023image} & 81.4 & 7.4 & 38.3 & 7.3 & 51.1 & 81.5 & 1.1 & 37.7 & 60.1 & 46.0 \\
        \bottomrule
        \end{tabular}
    }	
    }
    \caption{\textbf{Obstacle detection scores}. 
    The primary metrics are Average Precision (AP$\uparrow$) for pixel classification 
    and Average $F_1$ ($\overline{F_1}\uparrow$) for instance level detection. The evaluation protocol and metrics follow the {\it Segment Me If You Can} benchmark~\cite{Chan21b}. The upper section of the table contains benchmark results of methods that do not train specifically for the detection of road obstacles. In the lower part we list methods which do some kind of specific training.
    }
    \label{table:benchmarks_baselines_full}
    \end{table*}

\subsection{Qualitative examples}\label{sup:qualitative_examp}

\begin{figure*} 

\centering
\newcommand{\svgwidth}{\textwidth}
\tiny
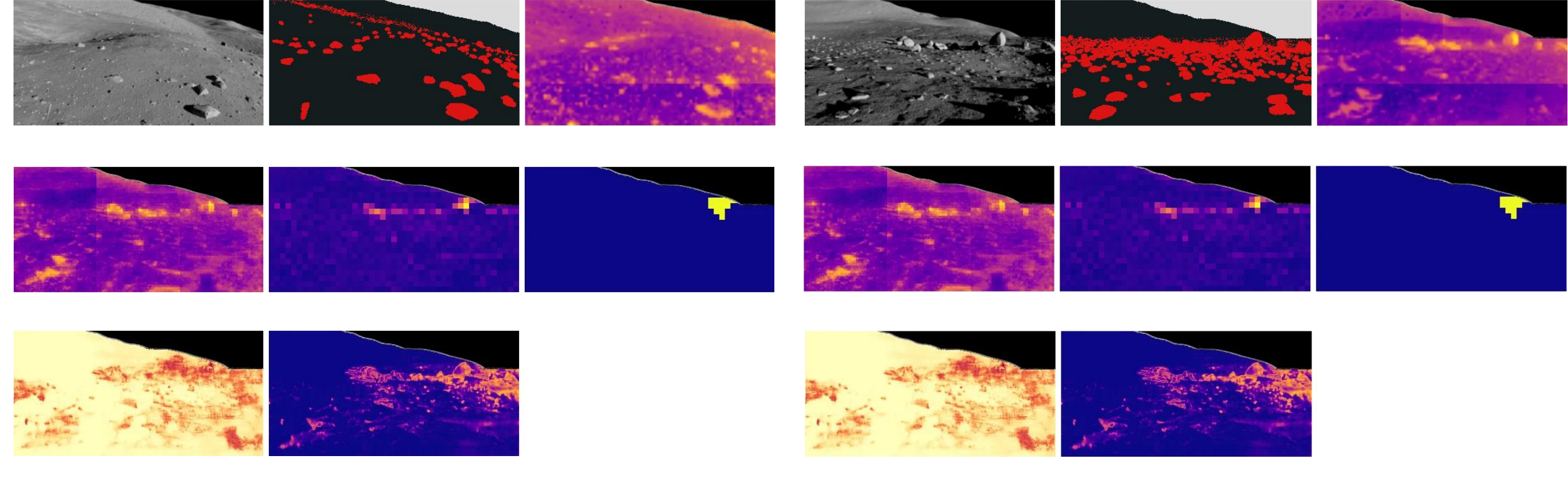
\caption{Qualitative results on moon rock detection on Artificial Lunar Landscape~\cite{ArtificialLunarLandscape}.
}
\label{fig:sup_quali_moon}
\end{figure*}

\begin{figure*} 
\centering
%
% {\tiny
% \includegraphics[width=0.48\linewidth]{images/sup_quali_obs/validation_5_comp.jpg}
% \includegraphics[width=0.48\linewidth]{images/sup_quali_obs/validation_18_comp.jpg}
% }
\newcommand{\svgwidth}{\textwidth}
\scriptsize
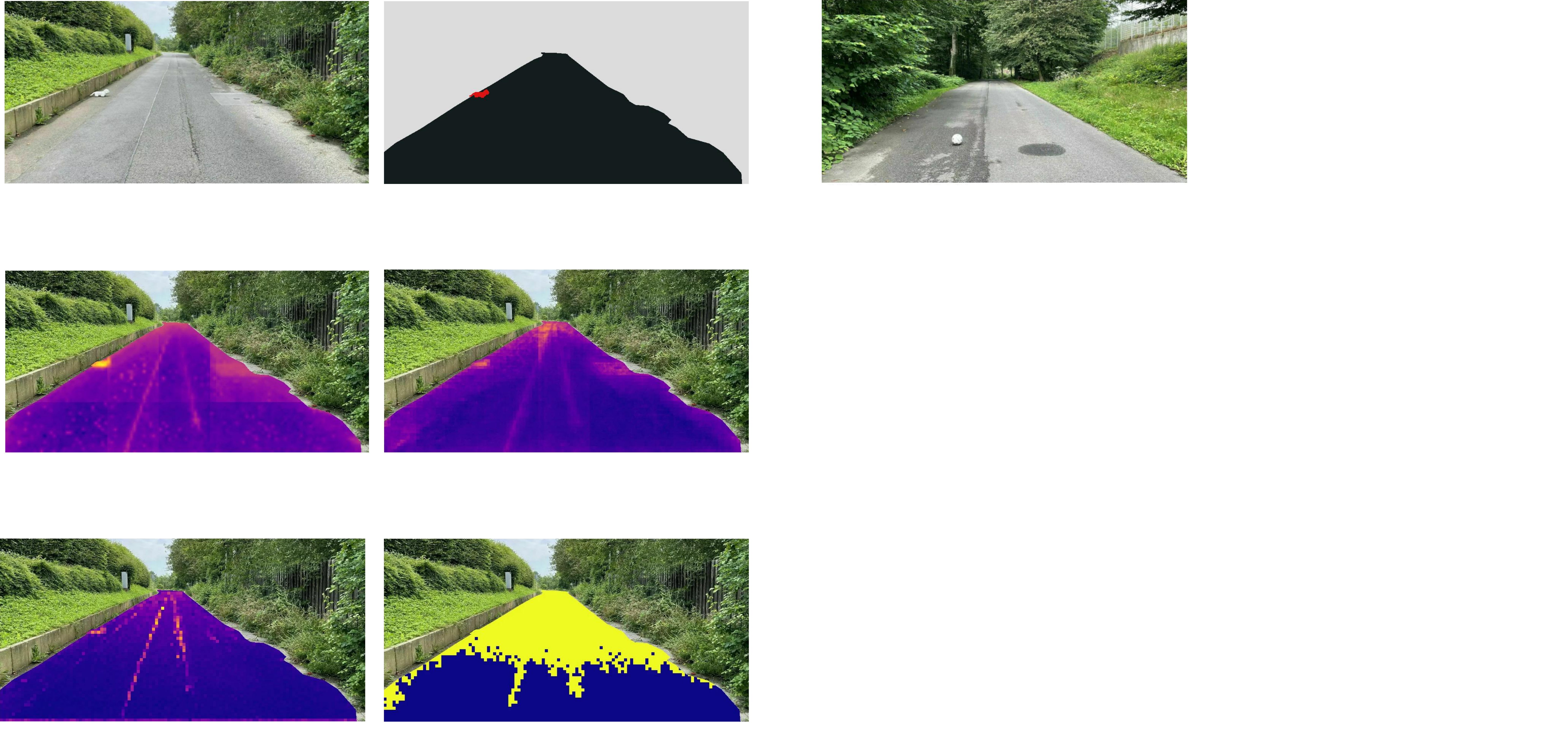
\caption{Qualitative results on obstacle detection on RoadObstacles21~\cite{Chan21b}.
}
\label{fig:sup_quali_obs}
\end{figure*}

\begin{figure*} 
\centering
\newcommand{\svgwidth}{\textwidth}
\tiny
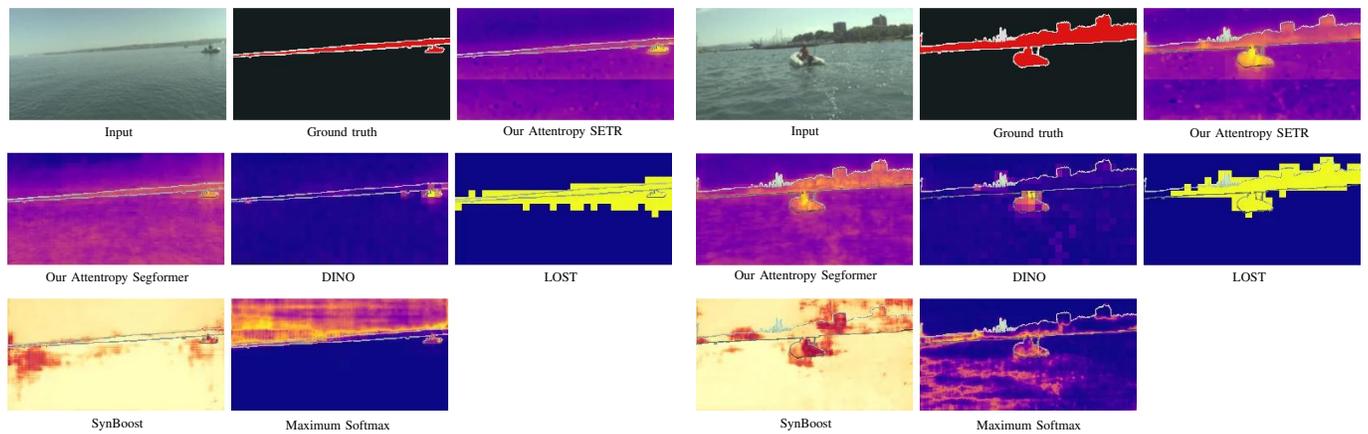
\caption{Qualitative results on obstacle detection on Maritime Dataset ~\cite{Bovcon19}.
}
\label{fig:sup_quali_mar}
\end{figure*}

In \cref{fig:sup_quali_moon}, \cref{fig:sup_quali_obs}, and \cref{fig:sup_quali_mar} we show extra qualitative comparisons.
For all datasets the entropy of the spatial attention layers of SETR leads to a meaningful segmentation of the small obstacles. This property generalizes to the Lunar (see \cref{fig:sup_quali_moon}) and Maritime (see \cref{fig:sup_quali_mar}) datasets where the transformer faces a complete domain shift. Compared to the training-based method SynBoost, our training-free AttEntropy method highlights the areas of the obstacles and moon rocks more clearly.

\end{document}